\documentclass[lettersize,journal]{IEEEtran}
\usepackage{amsmath,amsfonts}
\usepackage{algorithmic}
\usepackage{algorithm}
\usepackage{array}
\usepackage[caption=false,font=normalsize,labelfont=sf,textfont=sf]{subfig}
\usepackage{textcomp}
\usepackage{stfloats}
\usepackage{url}
\usepackage{verbatim}
\usepackage{graphicx}
\usepackage{cite}
\usepackage{multirow}
\usepackage[pdfborder={0 0 1}, linkbordercolor={1 0 0},  citebordercolor={0 1 0}, urlbordercolor={0 0 1} ]{hyperref}
\usepackage{orcidlink} 

\begin{document}

%~\IEEEmembership{Staff,~IEEE,}
\title{SAM2-ELNet: Label Enhancement and Automatic Annotation for Remote Sensing Segmentation}

\author{Jianhao Yang\,\orcidlink{0009-0006-4207-6347},~
	Wenshuo Yu\,\orcidlink{0000-0002-6943-6430},~ 
	Yuanchao Lv,~ 
	Jiance Sun,~  
	Bokang Sun,~
	and Mingyang Liu
    % <-this % stops a space
    
\thanks{This research was funded by The Science and Technology Department of Jilin Province of China under Grants 20240301024GX, and by Jilin University Innovation and Entrepreneurship Training Program(202410183293). (Corresponding authors: Mingyang Liu.)}
\thanks{Mingyang Liu, Jianhao Yang, Wenshuo Yu, Yuanchao Lv, and Bokang Sun are with College of Instrumentation and Electrical Engineering, Jilin University, Key Laboratory of Geophysical Exploration Equipment, Ministry of Education, Changchun 130000, China  (e-mail: liumingyang@jlu.edu.cn; yangjh1722@mails.jlu.edu.cn; yuwenshuo1998@163.com; lvyc23@mails.jlu.edu.cn; sunbk24@mails.jlu.edu.cn)}
\thanks{Jiance Sun is with the School of Mathematics, Jilin University, Changchun 130000, China  (e-mail: dd7159733@163.com. )}
}

\markboth{Journal of \LaTeX\ Class Files,~Vol.~00, No.~0, August~0000}%
{Yang \MakeLowercase{\textit{et al.}}: SAM2-ELNet: Label Enhancement and Automatic Annotation for Remote Sensing Segmentation}

\IEEEpubid{0000--0000/00\$00.00~\copyright~2025 IEEE}

\maketitle
\begin{abstract}
Remote sensing image segmentation is crucial for environmental monitoring, disaster assessment, and resource management, but its performance largely depends on the quality of the dataset. Although several high-quality datasets are broadly accessible, data scarcity remains for specialized tasks like marine oil spill segmentation. Such tasks still rely on manual annotation, which is both time-consuming and influenced by subjective human factors. The Segment Anything Model 2 (SAM2) has strong potential as an automatic annotation framework but struggles to perform effectively on heterogeneous, low-contrast remote sensing imagery. To address these challenges, we introduce a novel label enhancement and automatic annotation framework, termed SAM2-ELNet (Enhancement and Labeling Network). Specifically, we employ the frozen Hiera backbone from the pre-trained SAM2 as the encoder, while fine-tuning the adapter and decoder for different remote sensing tasks. Additionally, the proposed framework includes a label quality evaluator for filtering, ensuring the reliability of the generated labels. We design a series of experiments targeting resource-limited remote sensing tasks and evaluate our method on two datasets: the Deep-SAR Oil Spill (SOS) dataset with Synthetic Aperture Radar (SAR) imagery, and the CHN6-CUG Road dataset with Very High Resolution (VHR) optical imagery. The proposed framework can enhance coarse annotations and generate reliable training data under resource-limited conditions. Fine-tuned on only 30\% of the training data, it generates automatically labeled data. A model trained solely on these achieves slightly lower performance than using the full original annotations, while greatly reducing labeling costs and offering a practical solution for large-scale remote sensing interpretation.
\end{abstract}


\section{Introduction}
\IEEEPARstart{R}{emote} sensing typically refers to the technology of imaging the Earth system in a specific electromagnetic spectrum band on an aerospace or aviation platform to obtain multi-faceted characteristic information of the observed object. Its imaging methods \cite{ref1} include optical imaging, thermal infrared \cite{ref2}, hyperspectral \cite{ref3}, \cite{ref4}, and SAR \cite{ref5}.  Currently, various ways exist to extract valuable information from the massive amount of data. Remote sensing image segmentation has become a focal point of contemporary research due to its capability to delineate target regions and reveal terrestrial spatial structures.

Remote sensing image segmentation methods are mainly divided into traditional methods and deep learning-based methods. Traditional segmentation methods typically employ techniques such as thresholding, random forest \cite{ref37}, support vector machine \cite{ref38}, \cite{ref16}, and conditional random field \cite{ref17} to delineate different surface structure regions. When applied to simple scenes, these methods often yield satisfactory results. However, their sensitivity to radiometric noise, occlusive shadows, and topological ambiguities leads to issues, especially when dealing with high-dimensional features and complex spatial relationships.

\IEEEpubidadjcol

Driven by advancements in the field of computer vision (CV), deep learning techniques have become the leading approach for remote sensing image segmentation \cite{ref49}, \cite{ref50}, \cite{ref51}, \cite{ref52}, \cite{ref53}, \cite{ref67}. Supervised learning methods in this domain rely on large-scale, high-quality annotated datasets to effectively train robust models. However, most existing datasets are manually annotated. The inherent heterogeneity of targets in remote sensing images, coupled with noise interference, affects boundary clarity, and increases the difficulty of precise recognition. Due to the subjectivity of annotators and the inherent complexity of the task, annotators often generate annotations of suboptimal quality during the labeling process, which tend to omit details or over-simplify geometric shapes\cite{ref6}. As dataset scales expand exponentially, the corresponding annotation costs grow prohibitively. Therefore, reducing reliance on high-quality manual annotations, improving annotation efficiency and accuracy, and developing effective automated annotation frameworks are crucial for advancing intelligent remote sensing recognition and interpretation.

Interestingly, the computer vision community has encountered challenges similar to those in remote sensing. Research efforts are increasingly converging on strategies to alleviate the dependency on manually annotated training samples. Self-supervised learning (SSL) extracts meaningful patterns from large-scale unlabeled data, reducing annotation costs. Dosovitskiy et al. \cite{ref7} introduced the vision transformer (ViT), which brings the transformer architecture designed for natural language processing into visual tasks, leading to more efficient feature encoding. Meanwhile, contrastive learning \cite{ref9}, exemplified by methods like simple framework for contrastive learning of visual representations (SimCLR) \cite{ref11} and momentum contrast for unsupervised visual representation learning (MoCo) \cite{ref10}, refines feature representation by distinguishing between positive and negative sample pairs, leading to more stable and generalizable embeddings. Unlike contrastive learning methods, which rely on explicit positive and negative sample pairs, Grill et al. \cite{ref54} proposed bootstrap your own latent (BYOL), which adopts a momentum-based update strategy for the Target Network without requiring negative samples.  This enables contrast-free self-supervised learning and leads to more robust feature representations. In recent years, the emergence of large foundation models in computer vision has provided stronger support for downstream tasks. Caron et al. \cite{ref8} introduced  self-distillation with no labels (DINO), which leverages self-distillation with a student-teacher network to learn powerful visual representations without requiring labeled data. He et al. \cite{ref12} proposed masked autoencoders (MAE), which enhance visual representations by randomly masking image regions and training a decoder for reconstruction. Building on MAE’s success, the segment anything model (SAM) \cite{ref13} and its enhanced version, SAM2 \cite{ref14}, are pretrained on larger-scale datasets, endowing SAM2 with outstanding feature extraction capabilities.

With the growing adoption of these methods, researchers are seeking better ways to optimize manually annotated data while improving labeling efficiency in remote sensing applications. Compared with CV tasks, remote sensing images exhibit more severe noise and scale variations, posing greater challenges for end-to-end self-supervised methods. Wang et al. \cite{ref58} proposed an automatic annotation tool that integrates multiple deep learning models (including SAM), called X-Anylabeling. Its application to remote sensing image annotation still requires manual input, and its effectiveness remains limited in scenarios involving small object detection, high-noise images, or abnormal data distributions. To enhance the applicability of SAM in remote sensing domain, Luo et al. \cite{ref36} proposed the SAM-RSIS framework, providing important insights for subsequent research. Ma et al. \cite{ref35} proposed the SGO-SGB framework by incorporating object consistency loss and boundary preservation loss, addressing SAM’s fragmentation and boundary inaccuracy issues in semantic segmentation of remote sensing imagery. Chen et al. \cite{ref59} developed the RSPrompter method, which learns to generate category-specific prompts to enhance SAM’s applicability in remote sensing instance segmentation. Although SAM2 performs excellently in multiple tasks, its segmentation results often suffer from bias when applied to the automatic annotation process without manual prompts, and adding manual prompts significantly increases the cost. Additionally, these base models are typically trained on natural image data, making them difficult to directly apply to remote sensing data, which often features noise, heterogeneity, and low contrast. In practical applications, remote sensing data varies significantly across different scenarios, tasks (such as marine oil spill segmentation and road segmentation), and sensors (such as SAR and VHR). The performance of general large models in these diverse scenarios, especially in specific application areas, often falls short of ideal results.

To address this limitation, from a practical application perspective, we propose a novel label enhancement and automatic annotation framework, termed SAM2-ELNet (Enhancement and Labeling Network). Unlike semi-supervised or weakly supervised approaches that aim to improve model accuracy, this work focuses on building a data-centric solution for remote sensing segmentation in resource-constrained scenarios.  Specifically, we aim to improve data usability and quality through automatic annotation and label enhancement, providing a more reliable foundation for downstream tasks. The framework is designed to improve label quality in scenarios with large-scale coarse annotations and to enable high-quality automatic labeling in cases where only limited manual annotations are available. By freezing the backbone network of SAM2 as the encoder and fine-tuning the decoder and adapter for downstream tasks, we effectively address the domain adaptation challenge of SAM2 across different remote sensing data. Our framework also incorporates a label quality evaluator with error-correction feedback, where the high-quality reliable labels are further used for model fine-tuning, while lower-quality labels are re-labeled. Since the backbone is frozen, the framework allows for cost-effective, rapid fine-tuning with small datasets, customized applications in small-scale real-world scenarios. The proposed framework supports both ``low-cost few-shot annotation + large-scale automatic annotation" and ``low-cost coarse annotation + label enhancement", contributing to the advancement of large-scale remote sensing data processing and intelligent interpretation. The main contributions of this study are as follows:

\begin{enumerate}
	\item{To address the high data annotation costs and inconsistent quality in resource-constrained remote sensing applications, we propose a novel label enhancement and automatic annotation framework, SAM2-ELNet. This framework enables label enhancement for existing annotations and automatic annotation for unlabeled data under low-cost conditions.}
	\item{To generate more reliable labels, we design a label quality evaluator, enabling our framework to incorporate a feedback correction mechanism. To enhance the label structure representation of existing models, we adopt the Edge Attention Mechanism (EAM), which strengthens the model’s ability to capture fine boundaries.}
	\item{Starting with the question ``Can automatic annotation replace manual annotation?", we developed label quality and automatic annotation evaluation experiments tailored to remote sensing, offering insights for future research.}
\end{enumerate}

The remainder of this paper is organized as follows. Section II reviews related works on remote sensing image segmentation, automatic annotation methods, and the application of pretrained models in remote sensing. Section III presents the proposed SAM2-ELNet framework, providing a detailed description of its components. Section IV offers a comprehensive experimental analysis, including model performance evaluation, ablation studies, generalization capability assessment, and efficiency comparisons. Both qualitative and quantitative analyses are conducted to validate the effectiveness of the proposed framework. Finally, Section V concludes the study.

\section{Related Works}
\subsection{Advances in Annotation Strategies for Remote Sensing Image Segmentation}
In recent years, researchers have actively explored efficient annotation and training strategies to reduce dependence on high-quality manual annotations, lower costs, and enhance the accuracy of remote sensing image segmentation.

\begin{figure}[!t]
	\centering
	\includegraphics[width=\linewidth]{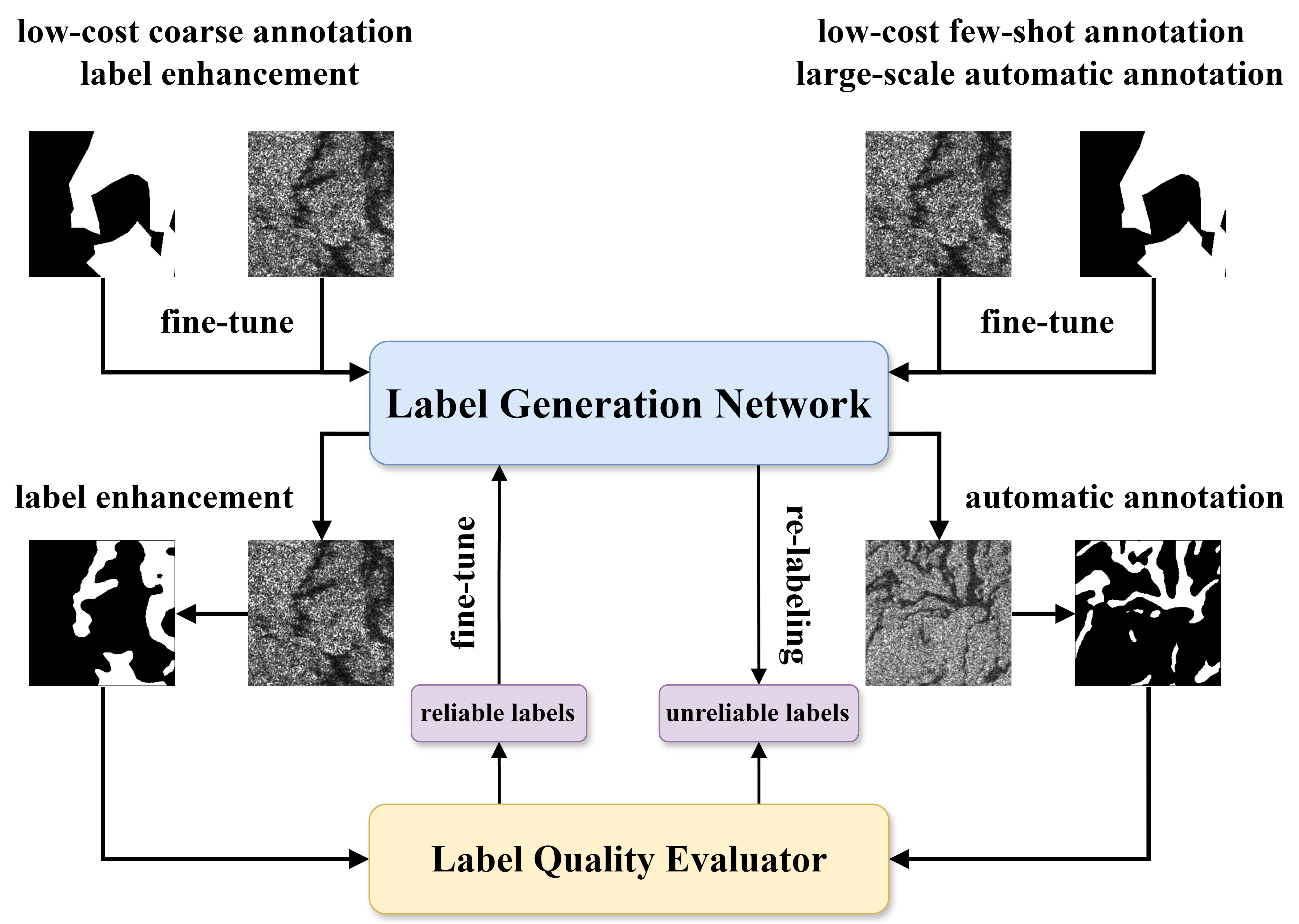}
	\caption{Architecture of SAM2-ELNet. The proposed SAM2-ELNet consists of a label generation network and a label quality evaluator. It enables label enhancement and automatic annotation in resource-constrained remote sensing segmentation scenarios through fine-tuning.}
	\label{fig_SAM2ELNet}
\end{figure}

Yang et al.\cite{ref18} introduced the EasySeg framework, which addresses domain adaptation challenges in remote sensing semantic segmentation.  Their method specifically targets domain discrepancies and misperception issues in cross-domain applications. It incorporates a 'See-First-Ask-Later' point-level annotation strategy and an Interactive Semantic Segmentation Network (ISS-Net), substantially reducing annotation costs while maintaining segmentation accuracy under weak supervision.  Liu et al.\cite{ref19} examined the role of noisy labels in supervised pretraining, analyzing their impact on encoder performance in remote sensing segmentation tasks. By comparing noisy label pretraining with self-supervised learning methods such as DINO and MoCo, they further demonstrated the effectiveness of noisy labels in pretraining and examined how class inconsistency influences model transferability. Yuan et al. \cite{ref63} combined MLLM with SAM to achieve more flexible and powerful segmentation functionality. Li et al.\cite{ref20} developed the SAM-OIL framework, which integrates YOLOv8-based object detection, an adapted SAM segmentation module, and an ordered mask fusion (OMF) mechanism for oil spill detection in SAR imagery. This method significantly improves detection accuracy while reducing dependence on fine-grained annotations.
\subsection{Enhancing Annotation Quality and Efficiency with Advanced Learning Models}
To further enhance annotation quality and efficiency,  Wu et al.\cite{ref21} pioneered the use of vision-language models (VLM) in farmland segmentation. They constructed a farmland image-text paired dataset and integrated semantic segmentation models with large language models (LLMs), developing the FSVLM model, which improved generalization and accuracy of remote sensing segmentation . Li et al.\cite{ref22} introduced a semi-automatic annotation method that combines anchor real-time lines and ROI real-time line algorithms to address label incompleteness and deficiencies in traditional remote sensing indices and interactive segmentation. Han et al.\cite{ref23} designed the DM-ProST framework, which generates high-quality pseudo-labels and expands training datasets through mutual correction between two deep learning models. This approach selects reliable unlabeled samples, improving segmentation accuracy. Zhu et al.\cite{ref24} proposed AIO2 (adaptively triggered online object label correction), which integrates adaptive correction triggering (ACT) and online object label correction (O2C) to enhance robustness against noisy labels. Shi et al.\cite{ref25} developed an automated multi-temporal change detection method for remote sensing images, incorporating boundary preservation techniques. By applying adaptive cropping and optimized classifier training, their method generates high-quality annotations and updates existing polygon-based labels. However, it does not resolve label omissions and struggles with non-polygonal targets such as oil spills and lakes.

\subsection{Optimization and Application of SAM in Remote Sensing and Geoscience}
The segment anything model has gained widespread attention for its role in self-supervised foundation models across remote sensing, medical imaging, and geophysical exploration\cite{ref26}. In land use and crop type segmentation, SAM significantly improves performance\cite{ref27}, \cite{ref28}.  Ma et al.\cite{ref29}  explored geophysical foundation models (GeoFM) for geophysical exploration and applied SAM to seismic phase picking. SAM also enhances building segmentation and urban area delineation in remote sensing imagery \cite{ref32}, \cite{ref33}, \cite{ref34}, \cite{ref68} demonstrating its adaptability in processing complex geographical features.
\begin{figure*}[!t]
	\centering
	\includegraphics[width=6.4in]{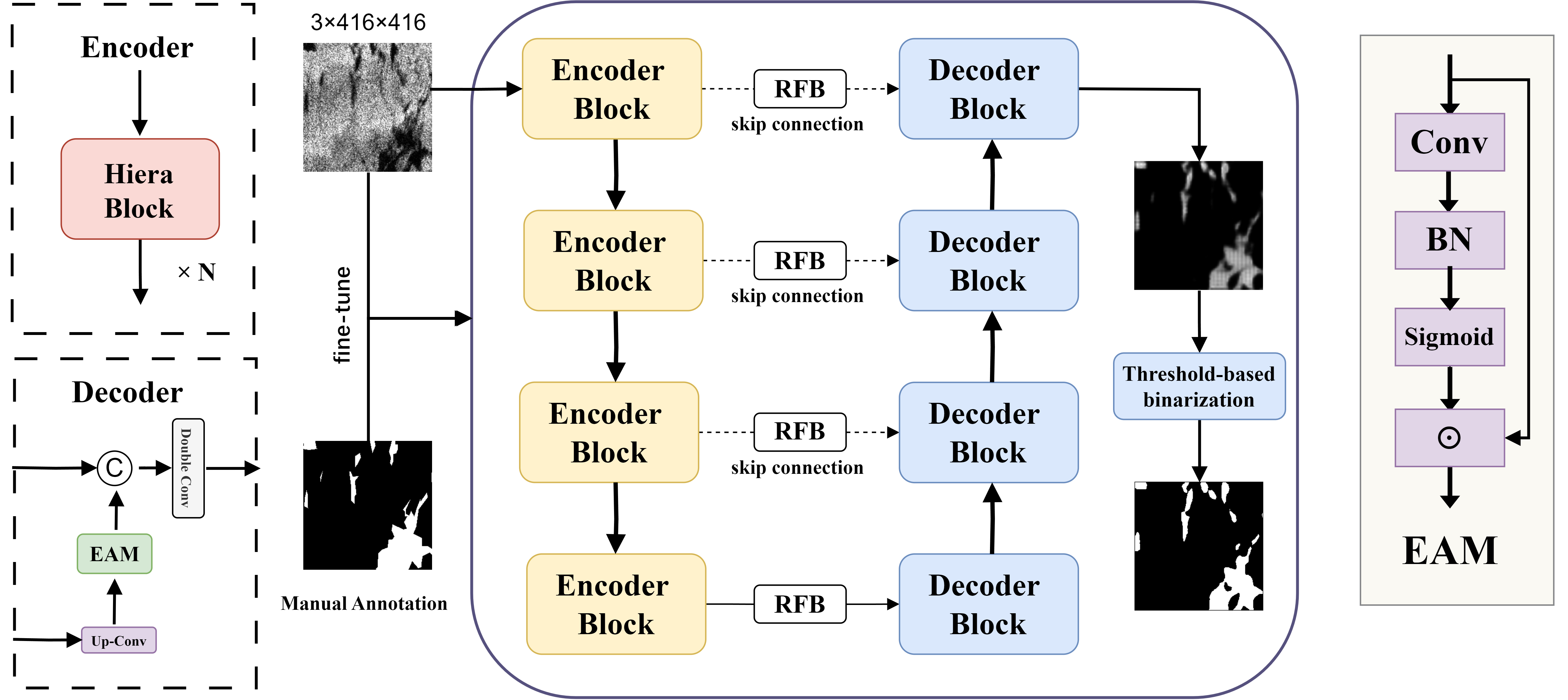}
	\caption{Architecture of the label generation network. The proposed network consists of multiple encoders, decoders, and receptive field blocks (RFB). It integrates EAM and parameter-efficient fine-tuning (PEFT) to enhance segmentation accuracy and improve label quality in remote sensing imagery.}
	\label{fig_labelgen}
\end{figure*}

\begin{figure}[!t]
	\centering
	\includegraphics[width=3.5in]{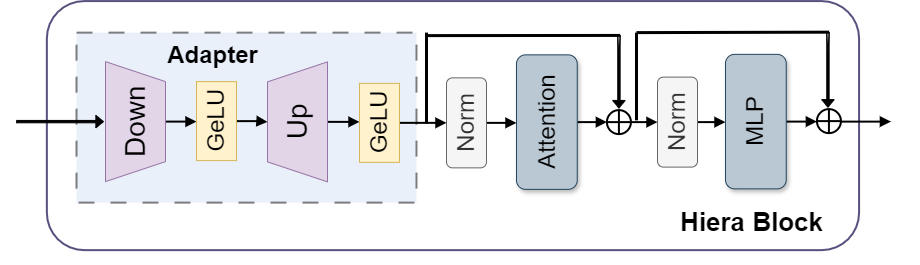} 
	\caption{Architecture of the Hiera. The encoder leverages the Hiera backbone for multi-scale feature modeling, with adapters enhancing feature adaptation.}
	\label{fig_h}
\end{figure}

\section{Proposed Method}
In this section, we introduce the basic framework of SAM2-ELNet. The proposed framework is designed for dataset construction and optimization in resource-constrained remote sensing segmentation scenarios. It revolves around two core modes: (1) low-cost few-shot annotation + large-scale automatic annotation, and (2) low-cost coarse annotation + label enhancement. As shown in Fig. \ref{fig_SAM2ELNet}, the framework mainly consists of two modules: a label generation network and a label quality evaluator for label filtering. The following subsections provide a detailed description of each module.

\subsection{Label Generation Network}
The label generation network is a pre-trained foundation model that can be rapidly fine-tuned for various practical remote sensing segmentation applications, particularly in resource-constrained scenarios. Building upon the advancements of SAM2-UNet \cite{ref15}, this paper proposes a label generation network specifically designed for annotation enhancement and automatic labeling. By integrating lightweight adapters, the network enables parameter-efficient fine-tuning (PEFT) while retaining the powerful feature extraction capabilities of the original pre-trained encoder.

As shown in Fig.\ref{fig_labelgen}, since remote sensing datasets are limited, retraining Hiera from scratch could weaken its feature extraction ability. To preserve its effectiveness, we adopt a frozen backbone strategy during training. Remote sensing images exhibit higher levels of noise and heterogeneity. To enable efficient fine-tuning while maintaining adaptability to remote sensing tasks, SAM2-ELNet inherits the adapter design from SAM2-UNet. We integrate adapters \cite{ref60}, \cite{ref61} before the multi-scale modules of the Hiera encoder, allowing the model to dynamically recalibrate feature distributions and reducing the number of tunable parameters. SAM2-ELNet employs a frozen encoder strategy, its PEFT approach is similar to low-rank adaptation \cite{ref39} (LoRA). LoRA mainly targets the attention mechanism of Transformer structures, but SAM2-ELNet embeds its adapter module before the multi-scale modules of the Hiera encoder to adjust feature mapping and enhance model adaptability, making it more suitable for remote sensing label generation tasks.

Hiera (hierarchical transformer) \cite{ref40} is a Transformer backbone network designed for efficient visual feature extraction. Unlike traditional convolutional neural networks (CNNs), Hiera captures both local details and global semantics through hierarchical feature modeling. Its layered computation strategy reduces computational cost while maintaining strong representational capacity. Our model employs the SAM2-pretrained Hiera as encoder, utilizing its multi-scale feature learning to generate high-quality segmentation features. Four receptive field blocks (RFBs) \cite{ref55}, \cite{ref56} further process the extracted features, reducing the channel count to 64 to enhance lightweight feature representation. The structure of the Hiera is illustrated in Fig. \ref{fig_h}.

\begin{figure*}[!t]
	\centering
	\includegraphics[width=5.6in]{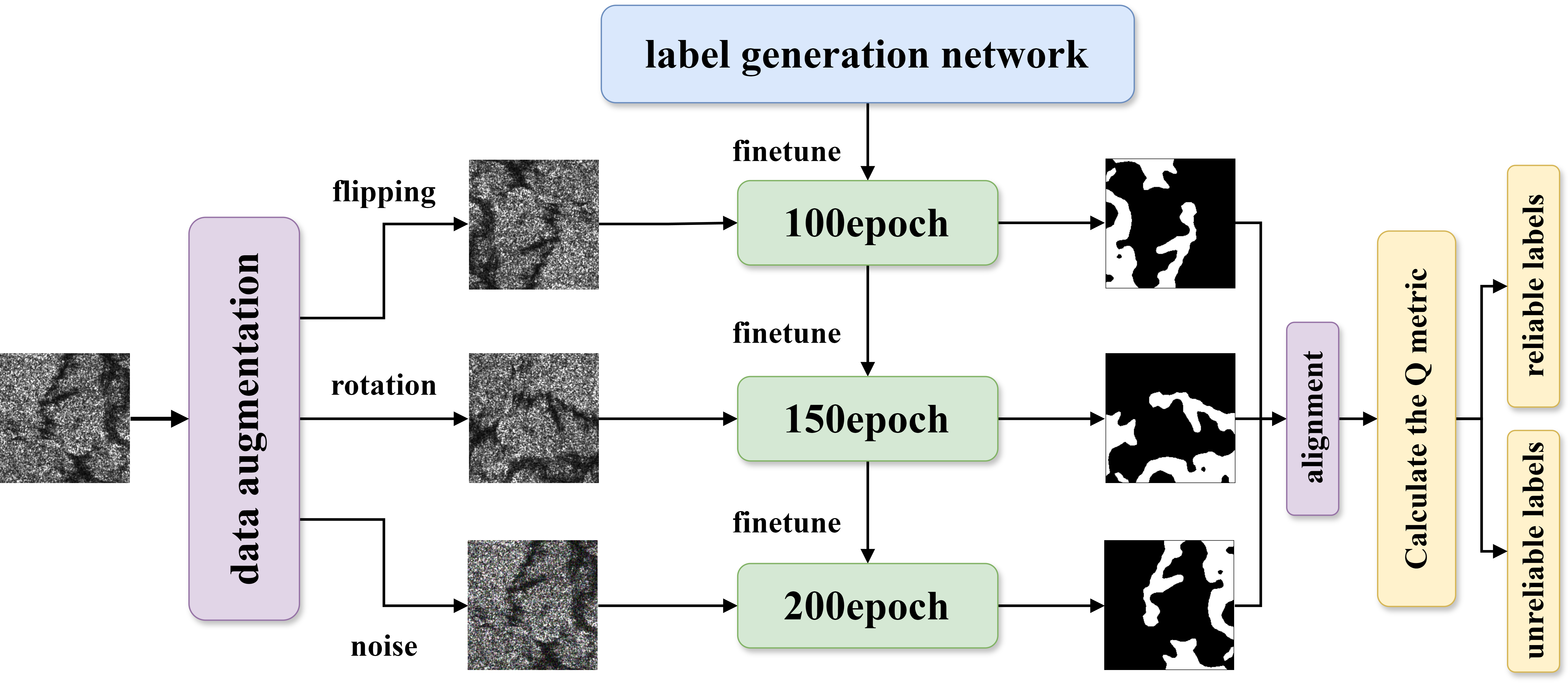} 
	\caption{Structure of the label quality evaluator. The module takes three perturbed inputs processed by differently fine-tuned label generation networks and evaluates prediction consistency. Low-quality labels are identified and flagged for re-annotation.}
	\label{fig_label}
\end{figure*}

Hiera’s large parameter count makes full fine-tuning computationally expensive. To enhance parameter efficiency, we adopt the SAM2-UNet design, freezing Hiera’s parameters and inserting lightweight adapters before each multi-scale block. Each adapter includes a downsampling linear layer followed by a GeLU \cite{ref57} activation, then an upsampling linear layer with another GeLU activation. This structure allows the model to efficiently recalibrate feature distributions while reducing the number of trainable parameters.

In remote sensing image segmentation tasks, accurately delineating object boundaries plays a critical role, especially when the target objects such as roads, marine oil spills, or man-made structures appear narrow, elongated, or fragmented. Inaccurate boundary handling can lead to serious consequences. For instance, boundary errors in road segmentation may disrupt connectivity analysis, affecting downstream applications like navigation or path planning. In oil spill detection, boundaries often lack clarity and exhibit smooth intensity transitions, making them hard to distinguish from the background. Without adequate boundary awareness, models tend to produce overly expanded prediction regions, significantly overestimating the affected area and misleading environmental assessments or emergency responses.

The baseline decoder follows the classical U-Net design, consisting of three decoding modules, each with two Conv-BN-ReLU layers. However, this structure struggles with blurred boundaries or low-contrast regions. To address this issue and enhance the model’s ability to capture fine boundary details, we introduce the EAM. Unlike traditional convolutional modules that aggregate spatial features uniformly, EAM focuses on the transition zones between foreground and background. It strengthens feature responses in edge regions and improves boundary awareness. The network applies EAM after each upsampling step to ensure that multi-scale feature fusion retains essential boundary cues. Specifically, EAM uses convolutional layers to extract edge-related features and applies element-wise weighting to refine the input features, guiding the model to emphasize edge information. The attention weights for edge refinement are computed as follows:

\begin{equation}
	\label{eam1}
	A_E = \sigma(\text{BN}(W_{\text{conv}} * X))
\end{equation}
where \( X \) is the input feature map, \( W_{\text{conv}} \) is a 3×3 convolution kernel. \( * \) denotes the convolution operation, BN refers to batch normalization, \( \sigma \) denotes the Sigmoid function.

The generated edge attention weight \( A_E \) modulates the original feature \( X \) to achieve edge enhancement:
\begin{equation}
	\label{eam2}
	X' = X \odot A_E
\end{equation}
where \( \odot \) represents element-wise multiplication. \( X' \) is an enhanced feature map emphasizing edge-related information.
This module adopts a lightweight design, using only an independent 3×3 convolutional layer, which significantly reduces computational cost compared to self-attention mechanisms.

\subsection{Loss Function}
We formulate the loss function as a combination of the weighted Intersection over Union (IoU) loss \( L_{\text{wIoU}} \) and the weighted binary cross-entropy (BCE) loss \( L_{\text{wBCE}} \), controlled by a balancing factor \( \lambda \), defined as:

\begin{equation}
	\label{eq_loss}
	L = \lambda \cdot L_{\text{wIoU}} + (1 - \lambda) \cdot L_{\text{wBCE}}
\end{equation}
To enhance training effectiveness, we apply deep supervision to all segmentation outputs \( S_i \), assigning each output a weight \( \alpha_i \). The final total loss function is defined as:

\begin{equation}
	\label{eq_total_loss}
	L_{\text{total}} = \sum_{i=1}^{3} \alpha_i \cdot L(G, S_i)
\end{equation}
where \( L(G, S_i) \) denotes the loss between the ground truth \( G \) and the predicted segmentation output \( S_i \), and \( \alpha_i \) is the supervision weight assigned to each level \( i \) of the segmentation output. The weights \( \alpha_i \) are normalized such that \( \sum \alpha_i = 1 \).

\subsection{Label Quality Evaluator}

Automatically generated labels inevitably contain noise and bias in downstream applications. To ensure label reliability, this study introduces a label quality evaluator that assesses the consistency of predicted labels and filters out unstable or low-confidence results. The structure of the evaluator is illustrated in Fig.~\ref{fig_label}.

Each remote sensing image \( I \) undergoes three types of basic data augmentation, such as Gaussian noise, horizontal flipping, and random rotation, resulting in three perturbed versions denoted as \( I_1, I_2, I_3 \). A label generation network with different fine-tuning epochs processes the three augmented inputs separately, producing segmentation predictions \( P_1, P_2, P_3 \). The three outputs undergo a mutual alignment process to ensure that all predictions share the same spatial orientation, enabling consistent pixel-wise comparison.

To quantify pixel-level similarity among the binarized predictions, the root mean squared error (RMSE) is used as the consistency metric. The RMSE score \( R_i \) at each pixel location \( i \) is calculated as:
\begin{equation}
	\label{eq_rmse}
	R_i = \sqrt{ \frac{1}{3} \sum_{j=1}^{3} \left(p_i^{(j)} - \bar{p}_i\right)^2 }
\end{equation}
In this formulation, \( p_i^{(j)} \in \{0, 1\} \) denotes the binary foreground prediction at pixel \( i \) in the \( j \)-th perturbed input. The mean prediction \( \bar{p}_i \) is computed as:
\begin{equation}
	\label{eq_mean_p}
	\bar{p}_i = \frac{1}{3} \sum_{j=1}^{3} p_i^{(j)}
\end{equation}
The RMSE reflects the level of disagreement at pixel \( i \). A lower \( R_i \) indicates stronger agreement among the three predictions, suggesting a more reliable label. To assess the overall consistency of a label, the final RMSE score is obtained by averaging \( R_i \) over all pixel positions.

Global consistency across binary segmentation masks is measured using the Intersection over Union (IoU) and Dice coefficient. For any two masks \( A \) and \( B \), these metrics are computed as:

\begin{equation}
	\label{eq_iou}
	\text{IoU}(A, B) = \frac{|A \cap B|}{|A \cup B|}
\end{equation}
\begin{equation}
	\label{eq_dice}
	\text{Dice}(A, B) = \frac{2|A \cap B|}{|A| + |B|}
\end{equation}
Average IoU and Dice scores across all three pairs of predictions \( P_1, P_2, P_3 \) are calculated using the following expressions:
\begin{equation}
	\label{eq_avg_iou}
	\text{IoU}_{\text{avg}} = \frac{1}{3} \sum_{1 \leq i < j \leq 3} \text{IoU}(P_i, P_j)
\end{equation}

\begin{equation}
	\label{eq_avg_dice}
	\text{Dice}_{\text{avg}} = \frac{1}{3} \sum_{1 \leq i < j \leq 3} \text{Dice}(P_i, P_j)
\end{equation}
A label is considered low-quality if the pixel consistency score \( C_i \), the average IoU, or the Dice score remains below a predefined threshold.

To obtain an overall label quality score, the pixel-wise RMSE score \( R \), average IoU, and average Dice coefficient are combined using a weighted formulation:

\begin{equation}
	\label{eq_final_score}
	Q = \beta_1 \cdot (1 - R) + \beta_2 \cdot \text{IoU}_{\text{avg}} + \beta_3 \cdot \text{Dice}_{\text{avg}}
\end{equation}
The weighting coefficients \( \beta_1, \beta_2, \beta_3 \) reflect the relative contribution of each metric. The RMSE term is transformed into \( 1 - R \) to ensure that a higher score consistently indicates better agreement among the predictions.

A label is marked as low-quality and scheduled for re-annotation if the final quality score \( Q \) falls below a predefined threshold.

\subsection{Overview of the Label Enhancement and Automatic Annotation Process}
The proposed label enhancement and automatic annotation process consists of the following stages:
\begin{enumerate}
	\item \textbf{Initial Label Preparation.} Prepare a small set of high-quality manually annotated samples when performing automatic annotation, or construct a large set of coarse manual annotations when conducting label enhancement. These datasets serve as the initial supervision sources for model fine-tuning.
	
	\item \textbf{Model Fine-Tuning.} Fine-tune the model using the prepared initial labels. The model initializes with pretrained weights, while only the adapters and decoder are updated during training to preserve the generalization ability of the encoder.
	
	\item \textbf{Automatic Label Generation and Filtering.} Apply the fine-tuned model to generate segmentation labels. The label quality evaluator filters out low-quality predictions, which are flagged for re-annotation to ensure label reliability.
	
	\item \textbf{Model Refinement with Filtered Labels.} Incorporate the retained high-quality labels into the training set and further fine-tune the model. This refined model can generate more accurate annotations in subsequent iterations.
	
	\item \textbf{Iterative Enhancement.} Repeat the label generation, quality filtering, and model refinement steps iteratively. This progressive process incrementally improves annotation quality and model performance in resource-constrained scenarios.
\end{enumerate}

\section{Experiments and Discussion}
\subsection{Datasets}
\subsubsection{Deep-SAR Oil Spill (SOS) dataset}
The Deep-SAR Oil Spill (SOS) dataset consists of SAR images from two sources \cite{ref46}: PALSAR data from the Mexican oil spill region and Sentinel data from the Persian Gulf. PALSAR, an L-band SAR sensor on the ALOS satellite, and Sentinel, a C-band SAR sensor, both provide all-weather, cloud-free imaging. In this dataset, dark regions in SAR images typically indicate oil spill contamination. Zhu et al.\cite{ref46} constructed the dataset from 21 SAR images, expanding it through cropping, rotation, and noise addition to produce 6,456 oil spill images \(416 \times 416\). The Mexican oil spill subset includes 3,101 training and 776 test images, while the Persian Gulf subset contains 3,354 training and 839 test images. Manual interpretation and GIS expert sampling generated the annotations, with optimizations addressing class imbalance.

\begin{figure}[!t]
	\centering
	\subfloat[Test sample 10011]{\includegraphics[width=1.6in]{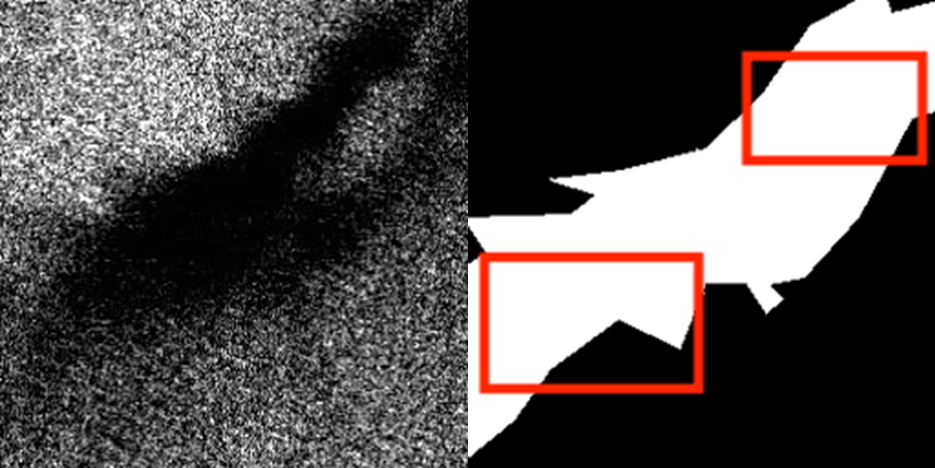}%
		\label{fig_pic1}}
	\hfil
	\subfloat[Test sample 10074]{\includegraphics[width=1.6in]{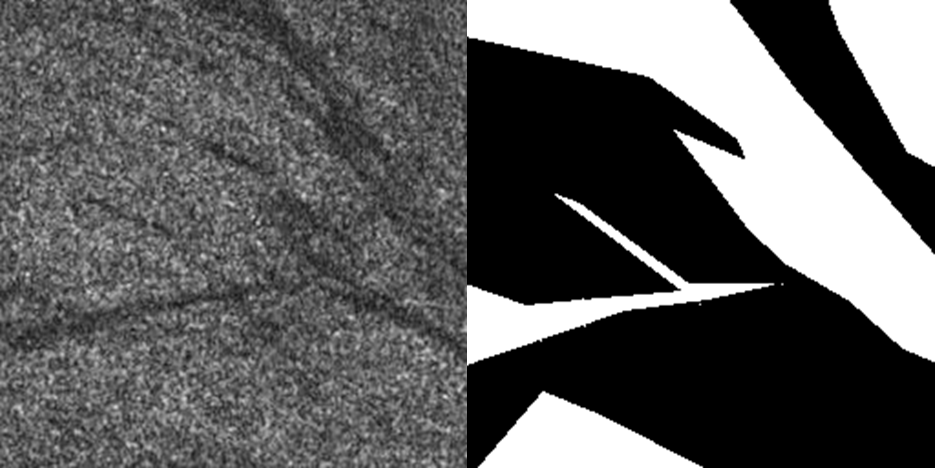}%
		\label{fig_pic2}}
	\hfil
	\subfloat[Train sample 10779]{\includegraphics[width=1.6in]{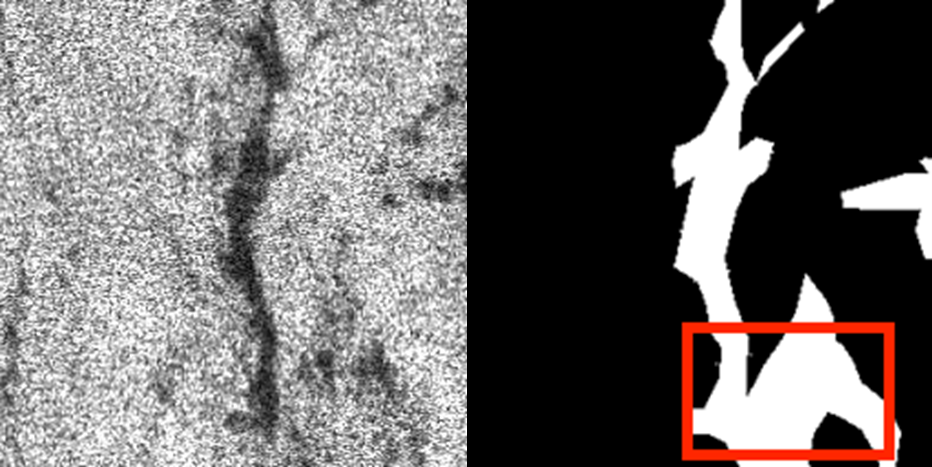}%
		\label{fig_pic3}}
	\hfil
	\subfloat[Train sample 11929]{\includegraphics[width=1.6in]{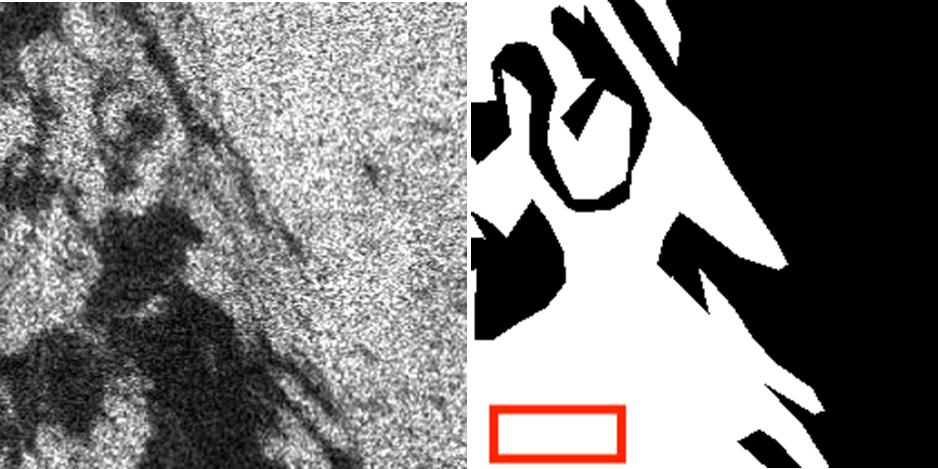}%
		\label{fig_pic4}}
	\caption{Examples of annotation inconsistencies in the SOS dataset. (a) The annotation erroneously classifies a sea region as an oil spill. (b) The annotated boundary exhibits an irregular polygonal shape, deviating from the original SAR image. (c) The label incorrectly expands a small oil spill into a large irregular polygon, while a portion of the oil spill on the left remains unlabeled. (d) The annotation misclassifies a sea region as an oil spill and exhibits irregular polygonal boundaries.}
	\label{fig_annotation_issues}
	
\end{figure}

Ideally, oil spill regions should have smooth and continuous boundaries. Due to the inherent subjectivity of manual annotation, as well as the heterogeneity of remote sensing images and noise interference, annotation results often exhibit issues such as missing label details, fragmentation, and polygonal boundaries. Illustrative examples are shown in Fig. \ref{fig_annotation_issues}.
\subsubsection{CHN6-CUG Road dataset}
The CHN6-CUG Road dataset \cite{ref64} is a high-resolution optical road segmentation dataset constructed from Google Earth imagery. It covers six representative cities in China with varying urban characteristics: Beijing (Chaoyang), Shanghai (Yangpu), Wuhan, Shenzhen (Nanshan), Hong Kong (Sha Tin), and Macao. The dataset includes 4,511 manually annotated images with a resolution of 50~cm/pixel and size \(512 \times 512\), of which 3,608 are used for training and 903 for testing. Labeled roads encompass various transportation types, including railways, highways, urban, and rural roads. Due to occlusions and complex urban structures, the annotations may exhibit challenges such as partial coverage and blurred boundaries, making this dataset suitable for evaluating segmentation robustness under realistic conditions.

\subsection{Hyperparameter Settings}
The experiment runs on an NVIDIA GeForce RTX 4090 GPU with 24GB of memory and utilizes CUDA 12.4 for accelerated computation. The model fine-tunes using the pretrained weights from sam2\_hiera\_large and optimizes with the Adam optimizer. The hyperparameters are set as follows: 200 training epochs, a batch size of 12, a learning rate of 0.001 and a weight decay of \(5 \times 10^{-4}\).

\subsection{Evaluation Metrics}
To assess the quality of segmentation results, we use Accuracy and Mean Intersection over Union (MIoU) as evaluation metrics.

\subsubsection{Overall Accuracy (ACC)}
Overall accuracy measures the model’s ability to correctly classify all categories. It is defined as:

\begin{equation}
	\text{ACC} = \frac{\text{TP} + \text{TN}}{\text{TP} + \text{FP} + \text{TN} + \text{FN}}
\end{equation}
where TP denotes the number of correctly predicted foreground pixels, TN represents the number of correctly predicted background pixels, FP refers to the number of pixels incorrectly classified as foreground, and FN indicates the number of pixels incorrectly classified as background.
\subsubsection{Mean Intersection over Union (mIoU)}
Mean Intersection over Union (mIoU) evaluates the overlap between the predicted segmentation and the ground truth. It is computed as:

\begin{equation}
	\text{mIoU} = \frac{1}{N} \sum_{c=1}^{N} \frac{\text{TP}_c}{\text{TP}_c + \text{FP}_c + \text{FN}_c}
\end{equation}
where TP\(_c\), FP\(_c\), and FN\(_c\) represent the true positive, false positive, false negative counts for class \( c \), and \( N \) is the total number of classes.

\subsection{Comparative Analysis of Label Enhancement}
\subsubsection{Qualitative Analysis}
To simulate the practical application scenario of ``low-cost coarse annotation + label enhancement" we treat the label annotations in the Deep-SAR Oil Spill (SOS) dataset as coarse labels to be enhanced. Both quantitative and qualitative evaluations are conducted accordingly.

The model is fine-tuned using only the training subset of the PALSAR dataset for 200 epochs. After training, it enhances both the training and test set labels of the PALSAR dataset, generating high-quality annotations for subsequent analysis.

\begin{figure*}[!t]
	\centering
	\captionsetup[subfloat]{labelformat=empty, position=top}
	\subfloat[\fontsize{9pt}{0pt}\selectfont original image]{\includegraphics[width=0.9in]{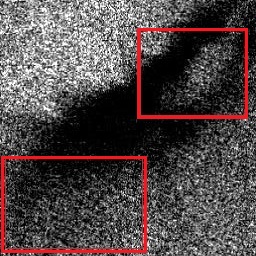}%
		\label{fig_pic11}}
	\hfil
	\subfloat[\fontsize{9pt}{0pt}\selectfont manual \vspace{2pt}]{\includegraphics[width=0.9in]{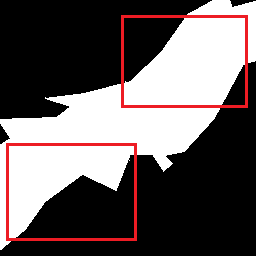}%
		\label{fig_pic12}}
	\hfil
	\subfloat[\fontsize{9pt}{0pt}\selectfont U-Net \vspace{2pt}]{\includegraphics[width=0.9in]{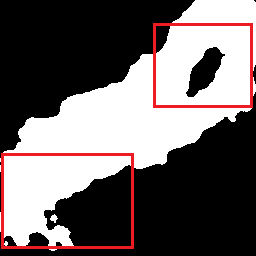}%
		\label{fig_pic13}}
	\hfil
	\subfloat[\fontsize{9pt}{0pt}\selectfont SAM2(Guidance) ]{\includegraphics[width=0.9in]{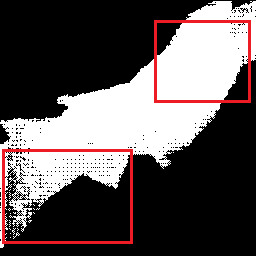}%
		\label{fig_pic14}}
	\hfil
	\subfloat[\fontsize{9pt}{0pt}\selectfont SAM2-UNet \vspace{1pt}]{\includegraphics[width=0.9in]{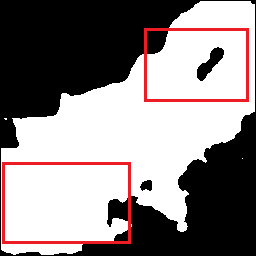}%
		\label{fig_pic15}}
	\hfil
	\subfloat[\fontsize{9pt}{0pt}\selectfont LOGCAN++ \vspace{1pt}]{\includegraphics[width=0.9in]{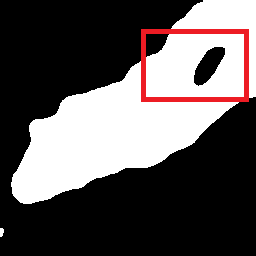}%
	\label{fig_pic17}}
	\hfil
	\subfloat[\fontsize{9pt}{0pt}\selectfont Ours(enhanced) ]{\includegraphics[width=0.9in]{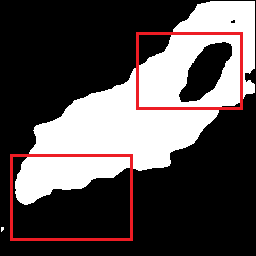}%
		\label{fig_pic16}}

	\vspace{-20pt}

	\subfloat[]{\includegraphics[width=0.9in]{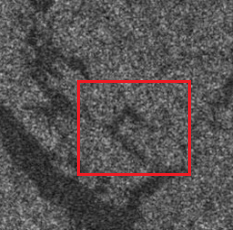}%
		\label{fig_pic21}}
	\hfil
	\subfloat[]{\includegraphics[width=0.9in]{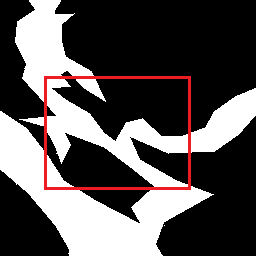}%
		\label{fig_pic22}}
	\hfil
	\subfloat[]{\includegraphics[width=0.9in]{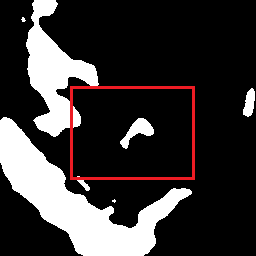}%
		\label{fig_pic23}}
	\hfil
	\subfloat[]{\includegraphics[width=0.9in]{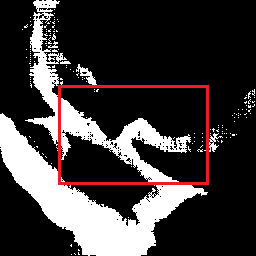}%
		\label{fig_pic24}}
	\hfil
	\subfloat[]{\includegraphics[width=0.9in]{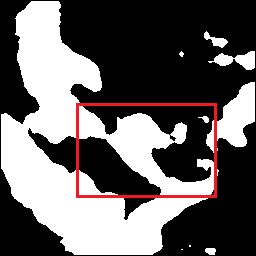}%
		\label{fig_pic25}}
	\hfil
	\subfloat[]{\includegraphics[width=0.9in]{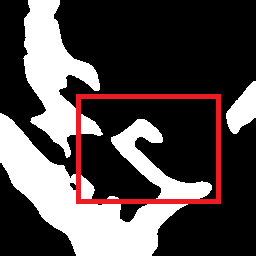}%
		\label{fig_pic27}}
	\hfil
	\subfloat[]{\includegraphics[width=0.9in]{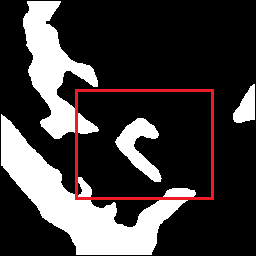}%
		\label{fig_pic26}}
	
	\vspace{-20pt}
	\subfloat[]{\includegraphics[width=0.9in]{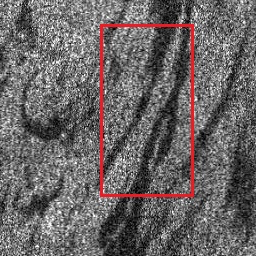}%
		\label{fig_pic31}}
	\hfil
	\subfloat[]{\includegraphics[width=0.9in]{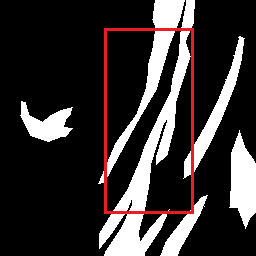}%
		\label{fig_pic32}}
	\hfil
	\subfloat[]{\includegraphics[width=0.9in]{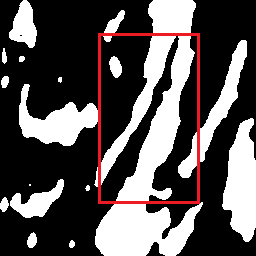}%
		\label{fig_pic33}}
	\hfil
	\subfloat[]{\includegraphics[width=0.9in]{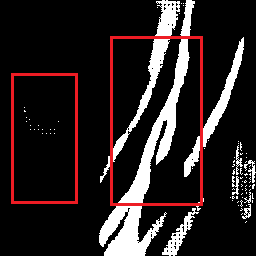}%
		\label{fig_pic34}}
	\hfil
	\subfloat[]{\includegraphics[width=0.9in]{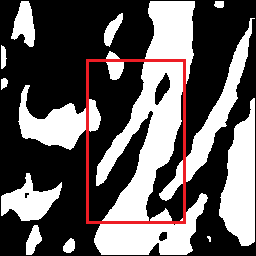}%
		\label{fig_pic35}}
	\hfil
	\subfloat[]{\includegraphics[width=0.9in]{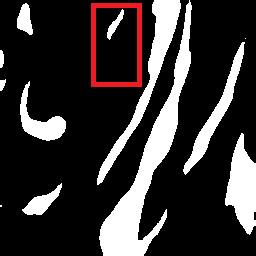}%
		\label{fig_pic37}}
	\hfil
	\subfloat[]{\includegraphics[width=0.9in]{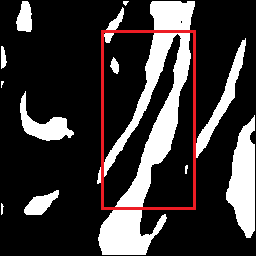}%
		\label{fig_pic36}}
	
	\vspace{-20pt}
	\subfloat[]{\includegraphics[width=0.9in]{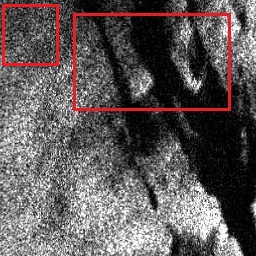}%
		\label{fig_pic41}}
	\hfil
	\subfloat[]{\includegraphics[width=0.9in]{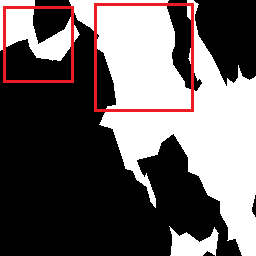}%
		\label{fig_pic42}}
	\hfil
	\subfloat[]{\includegraphics[width=0.9in]{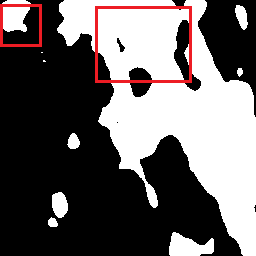}%
		\label{fig_pic43}}
	\hfil
	\subfloat[]{\includegraphics[width=0.9in]{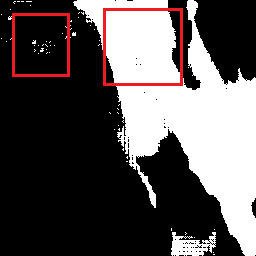}%
		\label{fig_pic44}}
	\hfil
	\subfloat[]{\includegraphics[width=0.9in]{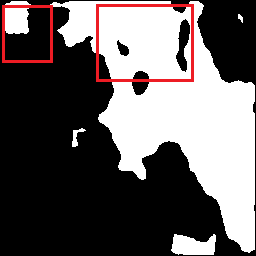}%
		\label{fig_pic45}}
	\hfil
	\subfloat[]{\includegraphics[width=0.9in]{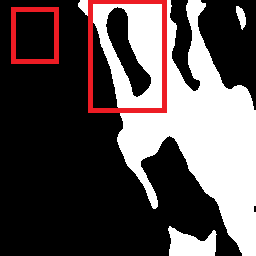}%
		\label{fig_pic47}}
	\hfil
	\subfloat[]{\includegraphics[width=0.9in]{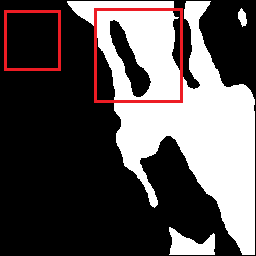}%
		\label{fig_pic46}}
	
	\vspace{-20pt}
	\subfloat[]{\includegraphics[width=0.9in]{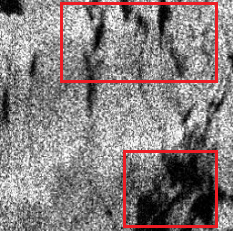}%
		\label{fig_pic51}}
	\hfil
	\subfloat[]{\includegraphics[width=0.9in]{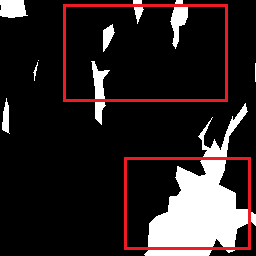}%
		\label{fig_pic52}}
	\hfil
	\subfloat[]{\includegraphics[width=0.9in]{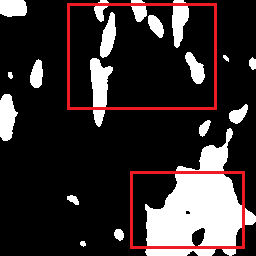}%
		\label{fig_pic53}}
	\hfil
	\subfloat[]{\includegraphics[width=0.9in]{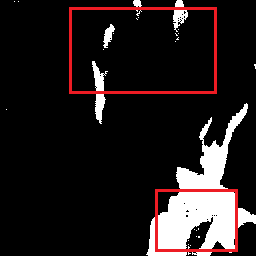}%
		\label{fig_pic54}}
	\hfil
	\subfloat[]{\includegraphics[width=0.9in]{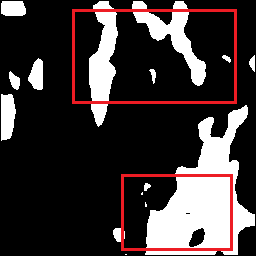}%
		\label{fig_pic55}}
	\hfil
	\subfloat[]{\includegraphics[width=0.9in]{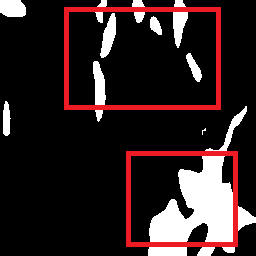}%
		\label{fig_pic57}}
	\hfil
	\subfloat[]{\includegraphics[width=0.9in]{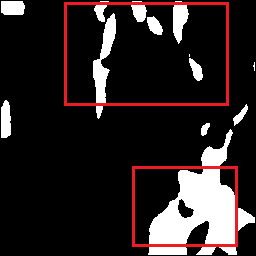}%
		\label{fig_pic56}}
	
	\caption{Comparison of label enhancement results across different methods.
		Each group shows the original SAR image, the manual annotation, and enhanced labels generated by our model, U-Net (same pipeline), SAM2 (guided by manual labels), SAM2-UNet, and LOGCAN++. Our method produces labels with finer detail, more accurate boundaries, and overall higher annotation quality, achieving comparable performance to current state-of-the-art approaches. }
	\label{fig_label_enhancement}
\end{figure*}

As shown in Fig. \ref{fig_label_enhancement}, the enhanced labels exhibit smoother and more continuous boundaries, and significantly refine annotation details compared to the original labels. This improvement appears consistently across the dataset. The enhanced annotations provide better object shape precision and more accurate representation of complex regions and ambiguous edges, enabling clearer delineation of target contours. These results demonstrate the model’s ability to effectively optimize manual annotations.

To further evaluate the model, we trained a U-Net using the same pipeline. U-Net \cite{ref62} showed moderate improvements, but our model produced more precise and consistent annotations. We also evaluated SAM2 \cite{ref14} in two variants: SAM2 (guided by manual labels) and SAM2-UNet. Compared to SAM2 (guided by manual labels), our model achieved noticeably better boundaries and structure. Compared to SAM2-UNet, it provided slightly finer details and more stable quality. In addition, we compared our approach with the current SOTA segmentation model LOGCAN++ \cite{ref66}, and achieved comparable performance, further demonstrating the effectiveness of our method in label quality enhancement tasks.

\subsubsection{Quantitative Analysis}
Our objective is to generate higher-quality annotations. Directly evaluating models using only the original labels may fail to fully reflect the effectiveness of the enhanced annotations. Moreover, since segmentation models tend to exhibit low sensitivity to label quality during training, the presence of noisy annotations may even improve model robustness. To address these limitations, we design a new evaluation strategy that constructs multiple test sets to more accurately assess annotation quality.

\begin{table*}[!t]
	\caption{Label Quality Evaluation Using Third-Party Segmentation Models}
	\label{tab:label_eval}
	\centering
	\renewcommand{\arraystretch}{1.2}
	\begin{tabular}{c c c c c c c}
		\hline
		\textbf{Test Model} & \textbf{Train Set} & \textbf{Test Set} & \textbf{mIoU} & \textbf{\(\Delta\) from HQ (mIoU)} & \textbf{ACC} & \textbf{\(\Delta\) from HQ (ACC)} \\
		\hline
		\multirow{3}{*}{DeepLabv3+} & \multirow{3}{*}{Original} 
		& Test-HQ & 0.7815 & - & 0.9283 & - \\
		& & Test-Orig & 0.7541 & 0.0274 & 0.9157 & 0.0127 \\
		& & Test-Enh  & \textbf{0.7877} & \textbf{0.0062} & \textbf{0.9328} & \textbf{0.0045} \\
		\hline
		\multirow{3}{*}{U-Net} & \multirow{3}{*}{Original}
		& Test-HQ & 0.8276 & - & 0.9462 & - \\
		& & Test-Orig & 0.7919 & 0.0357 & 0.9350 & 0.0112 \\
		& & Test-Enh  & \textbf{0.8369} & \textbf{0.0093} & \textbf{0.9491} & \textbf{0.0029} \\
		\hline
	\end{tabular}
\end{table*}

\begin{table*}[!t]
	\caption{Evaluation of Enhanced Labels Using Baseline Models}
	\label{tab:baseline_results}
	\centering
	\renewcommand{\arraystretch}{1.2}
	\begin{tabular}{c c c c c c c}
		\hline
		\textbf{Test Model} & \textbf{Train Set} & \textbf{Test Set} & \textbf{mIoU} & \textbf{\(\Delta\) from HQ (mIoU)} & \textbf{ACC} & \textbf{\(\Delta\) from HQ (ACC)} \\
		\hline
		\multirow{5}{*}{DeepLabv3+} & \multirow{5}{*}{Original}
		& Test-HQ & 0.7815 & - & 0.9283 & - \\
		& & Test-U-Net & 0.8517 & 0.0701 & 0.9466 & 0.0182 \\
		& & Test-SAM2-UNet & 0.8192 & 0.0377 & 0.9360 & 0.0076 \\
		& & Test-LOGCAN++ & 0.7893 & 0.0077 & 0.9324 & \textbf{0.0041} \\
		& & Test-Ours & 0.7877 & \textbf{0.0062} & 0.9328 & 0.0045 \\
		\hline
	\end{tabular}
\end{table*}

Specifically, a group of domain experts were invited to refine the annotations of selected samples in the test set, forming a high-quality benchmark test set referred to as Test-HQ. Corresponding samples from the original and enhanced labels were also selected to construct two comparison test sets, Test-Orig and Test-Enh. A third-party segmentation model undergoes training with the original training set and evaluation on the three test sets. Performance metrics offer an indirect means to evaluate the actual quality of both original and enhanced labels. Although traditional metrics may not fully reflect the quality of annotations, the performance gap between these results and Test-HQ enables a more objective assessment of label quality.

\begin{figure*}[!t]
	\centering
	\captionsetup[subfloat]{labelformat=empty}
	\subfloat[Loss]{\includegraphics[width=2.2in]{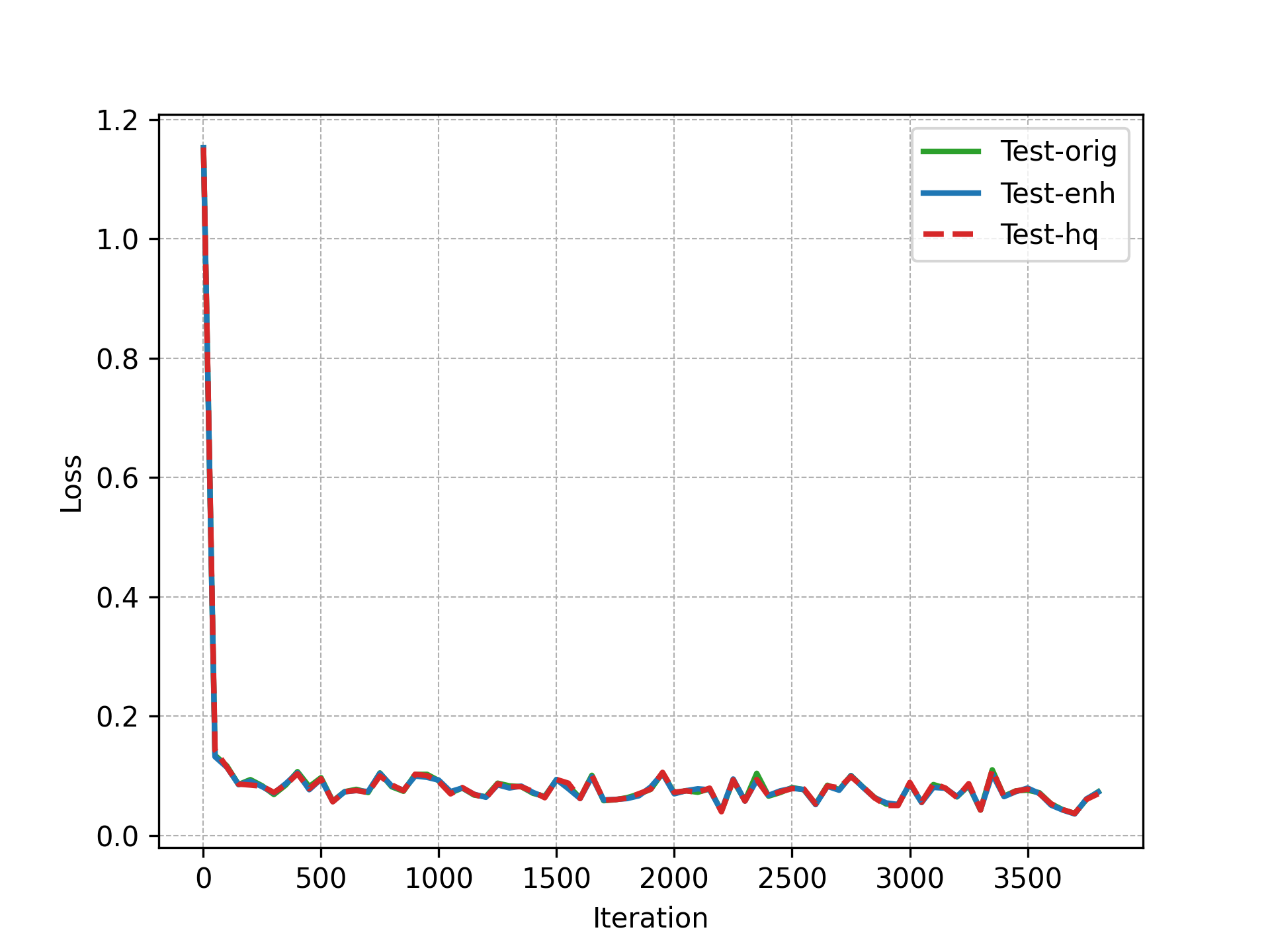}\label{fig_loss_enh}}
	\hfil
	\subfloat[mIoU]{\includegraphics[width=2.2in]{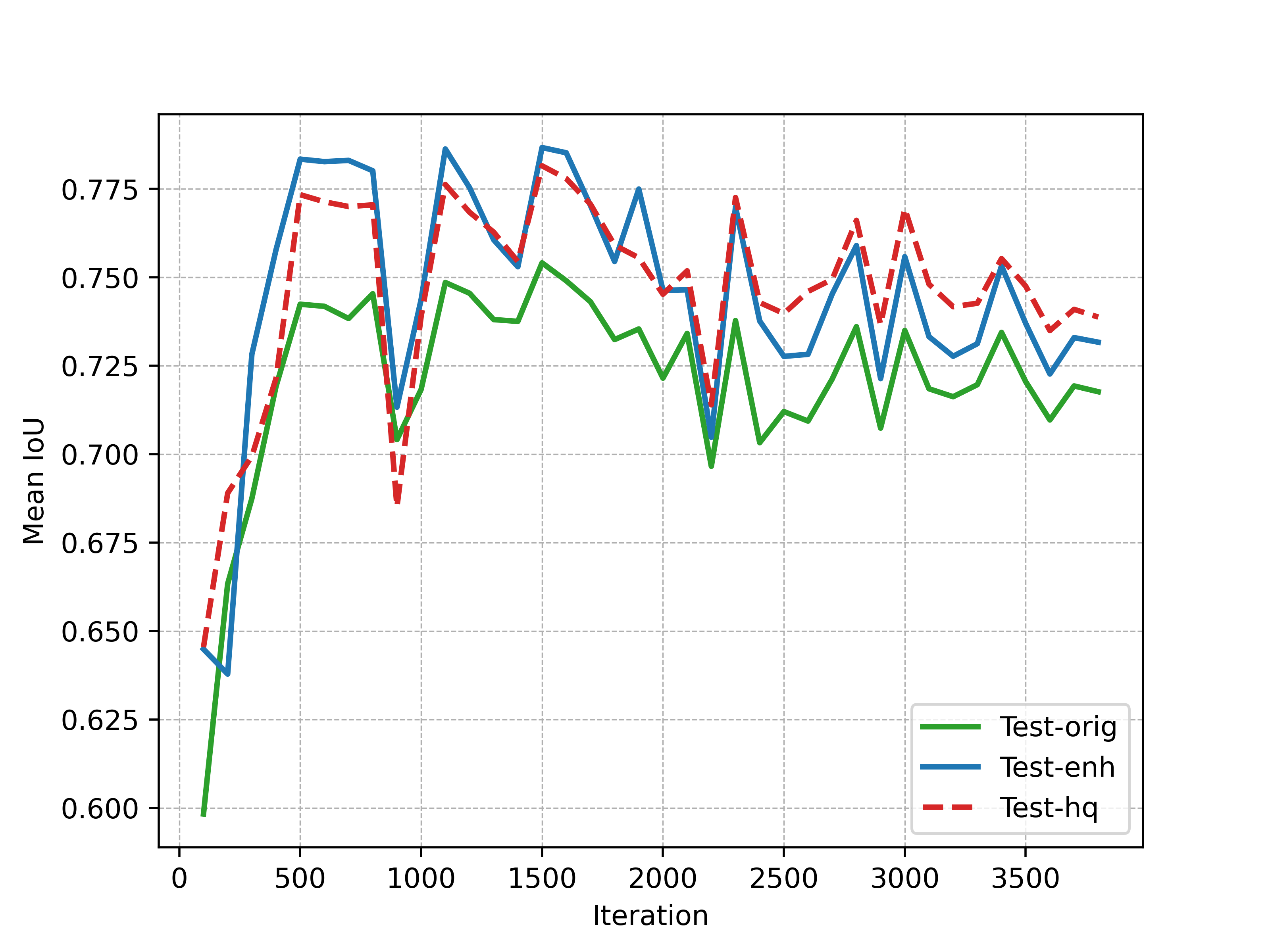}\label{fig_miou_enh}}
	\hfil
	\subfloat[Accuracy]{\includegraphics[width=2.2in]{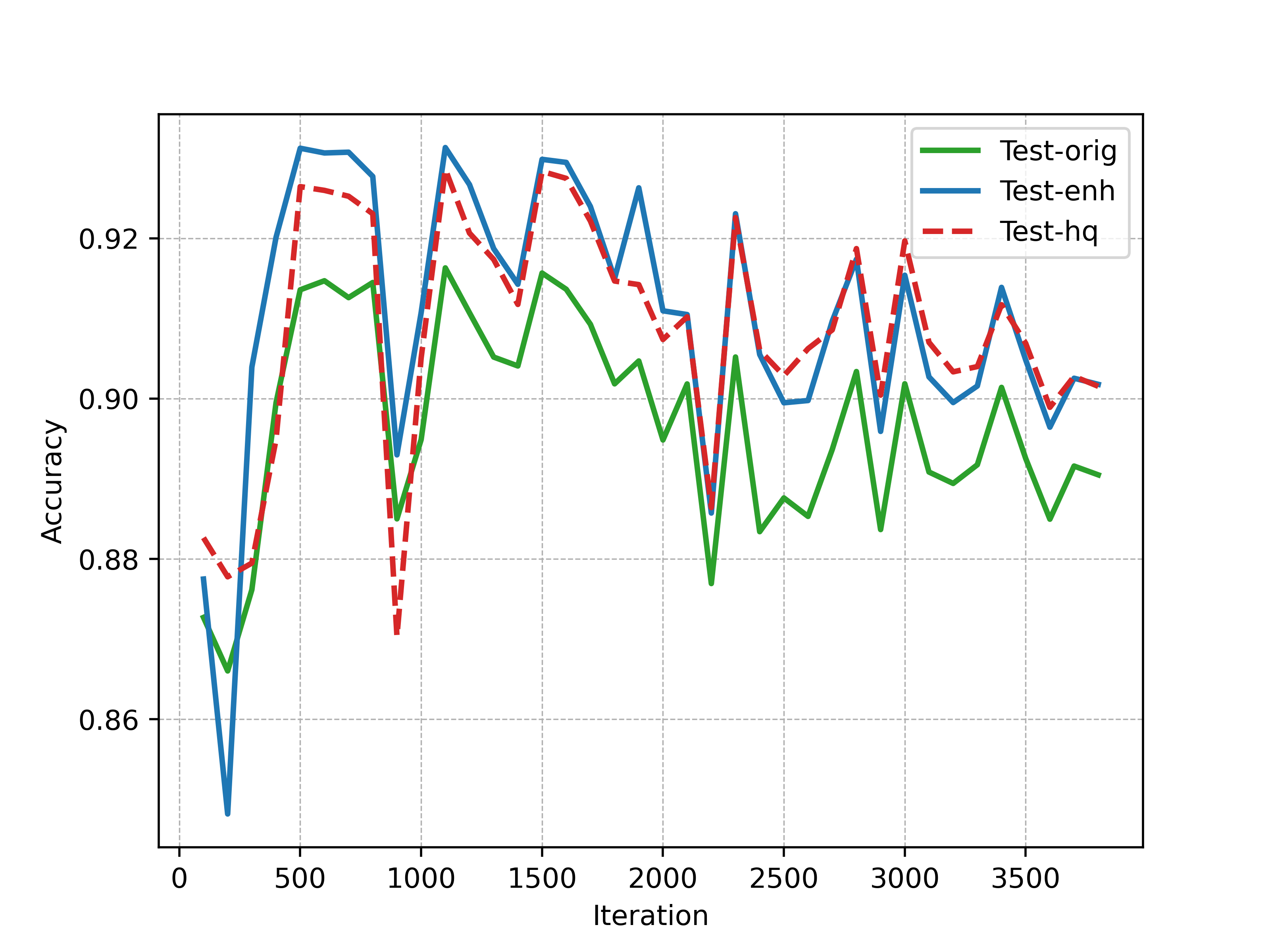}\label{fig_acc_enh}}
	
	\caption{Evaluation results of the model trained on original labels across three different test sets. The performance on Test-Enh is notably closer to that on Test-HQ, indicating the higher quality of the enhanced annotations.}
	\label{fig_enh}
\end{figure*}

\begin{figure*}[!t]
	\centering
	\captionsetup[subfloat]{labelformat=empty, position=top}
	
	\subfloat[\fontsize{9pt}{0pt}\selectfont original image]{\includegraphics[width=1.1in]{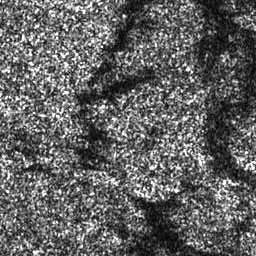}%
		\label{fig_pict11}}
	\hfil
	\subfloat[\fontsize{9pt}{0pt}\selectfont manual labels \vspace{2pt}]{\includegraphics[width=1.1in]{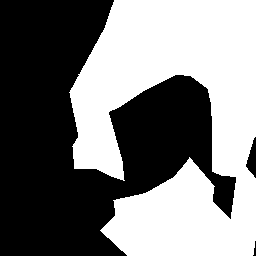}%
		\label{fig_pict12}}
	\hfil
	\subfloat[\fontsize{9pt}{0pt}\selectfont auto-labels \vspace{2pt}]{\includegraphics[width=1.1in]{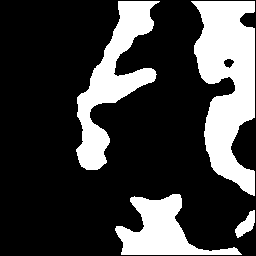}%
		\label{fig_pict13}}
	\hfil
	\subfloat[\fontsize{9pt}{0pt}\selectfont original image]{\includegraphics[width=1.1in]{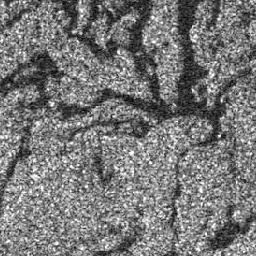}%
		\label{fig_pict21}}
	\hfil
	\subfloat[\fontsize{9pt}{1pt}\selectfont manual labels \vspace{2pt}]{\includegraphics[width=1.1in]{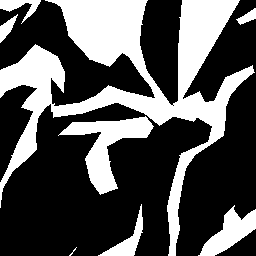}%
		\label{fig_pict22}}
	\hfil
	\subfloat[\fontsize{9pt}{1pt}\selectfont auto-labels \vspace{2pt}]{\includegraphics[width=1.1in]{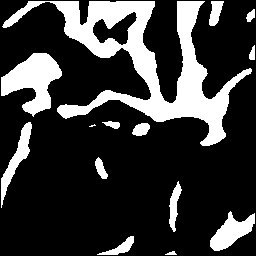}%
		\label{fig_pict23}}
	
	\vspace{-20pt}
	
	\subfloat[]{\includegraphics[width=1.1in]{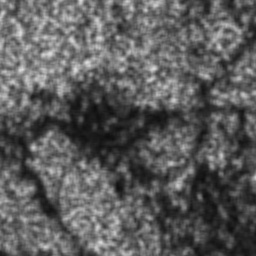}%
		\label{fig_pict31}}
	\hfil
	\subfloat[]{\includegraphics[width=1.1in]{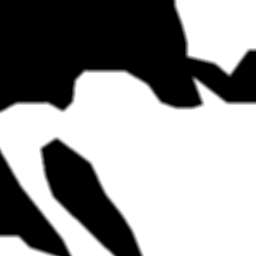}%
		\label{fig_pict32}}
	\hfil
	\subfloat[]{\includegraphics[width=1.1in]{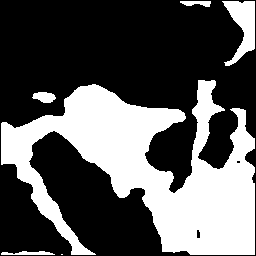}%
		\label{fig_pict33}}
	\hfil
	\subfloat[]{\includegraphics[width=1.1in]{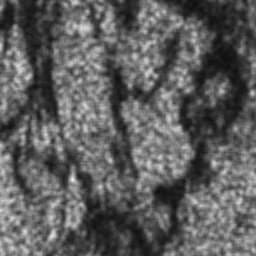}%
		\label{fig_pict41}}
	\hfil
	\subfloat[]{\includegraphics[width=1.1in]{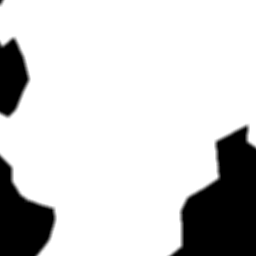}%
		\label{fig_pict42}}
	\hfil
	\subfloat[]{\includegraphics[width=1.1in]{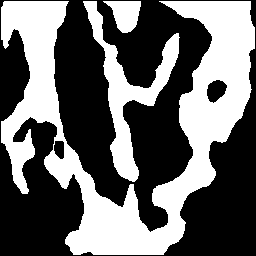}%
		\label{fig_pict43}}
	
	\caption{Generalization performance of automatic annotation on the Sentinel dataset. The automatically generated labels not only preserve key target structures but also exhibit improved quality in certain regions compared to the original annotations.}
	\label{fig_Generalization}
\end{figure*}

In this experiment, we adopt open-source implementations of DeepLabv3+ and U-Net \cite{ref46, ref47, ref62} to analyze the quality of the generated annotations. Specifically, the performance of models trained on original labels is evaluated across the three test sets, and the performance gap relative to Test-HQ is calculated to assess label quality more accurately. All experiments are conducted under the same training conditions. The backbone networks for DeepLabv3+ include MobileNet, ResNet50, and ResNet101. Each model is initialized with the same pretrained weights and trained for 20 epochs using consistent hyperparameters.

The experimental results in Table~\ref{tab:label_eval} and Fig.~\ref{fig_enh} show that models trained on the original labels perform significantly better on both Test-Enh and Test-HQ in terms of mean Intersection over Union (mIoU) and accuracy (ACC), compared to their performance on Test-Orig. Moreover, the results on Test-Enh are closely aligned with those on Test-HQ, further demonstrating that the enhanced labels have achieved a quality level comparable to expert-level fine annotations.

To verify the effectiveness of our method, we applied the same evaluation protocol to several representative baseline models. The results are summarized in Table~\ref{tab:baseline_results}.

When using model-generated labels as test sets, evaluation scores often exceed those obtained with manual annotations. This happens because coarse or overly smooth test labels make it easier for the model to produce matching predictions, thus inflating the evaluation metrics. Visual inspection reveals that labels generated by U-Net tend to cover broader regions. When used as test labels, they yield higher scores and exhibit a larger gap from the high-quality reference set (Test-HQ).

In contrast, our method consistently delivers results that are closer to Test-HQ, indicating that the generated labels better approximate expert-level fine annotations. Furthermore, our approach achieves comparable performance to the state-of-the-art model LOGCAN++, reinforcing the reliability of the proposed evaluation scheme and confirming the effectiveness of the label enhancement framework.

\begin{figure*}[!t]
	\centering
	\captionsetup[subfloat]{labelformat=empty}
	\subfloat[Loss]{\includegraphics[width=2.2in]{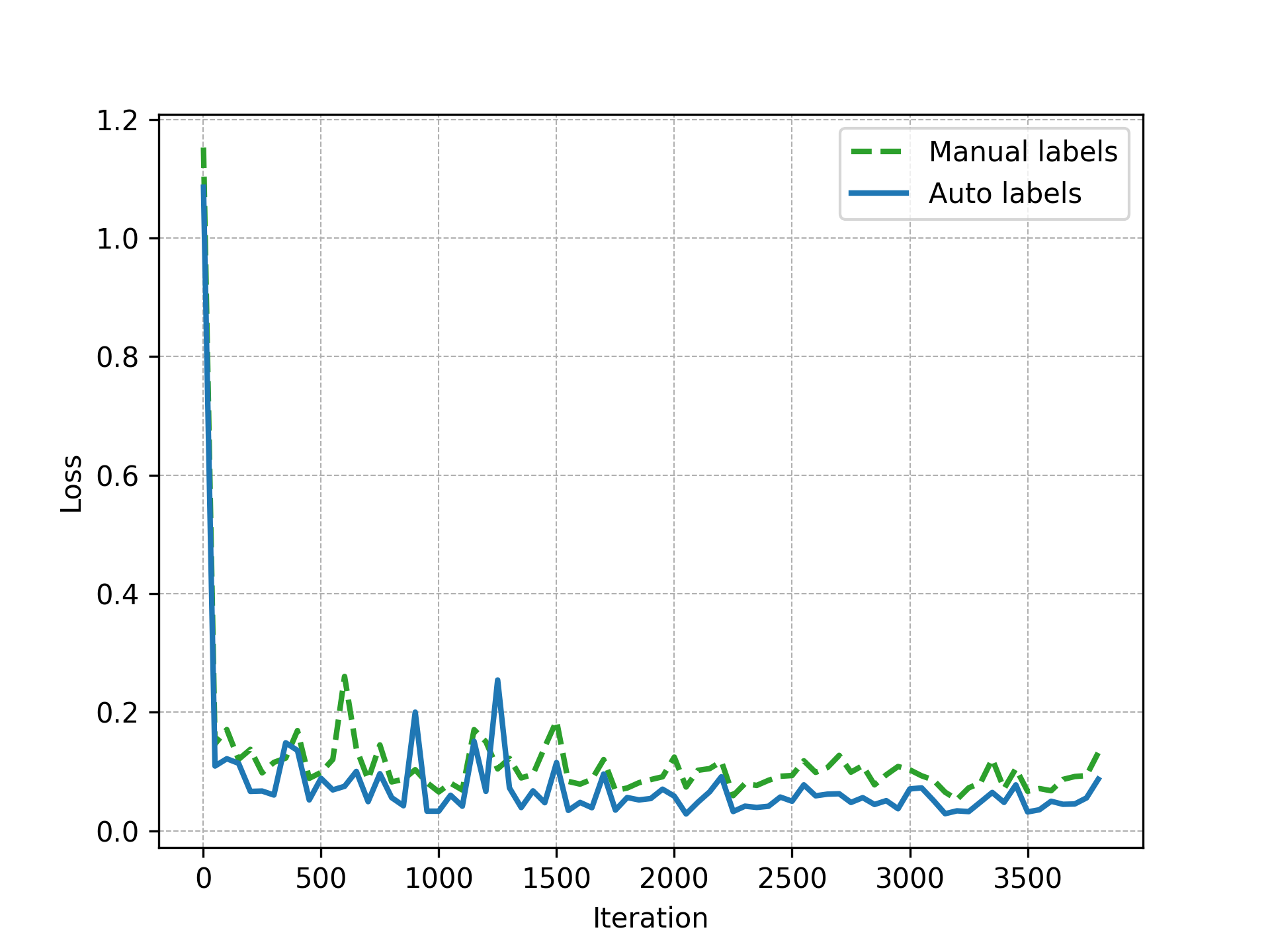}\label{fig_loss_gen}}
	\hfil
	\subfloat[mIoU]{\includegraphics[width=2.2in]{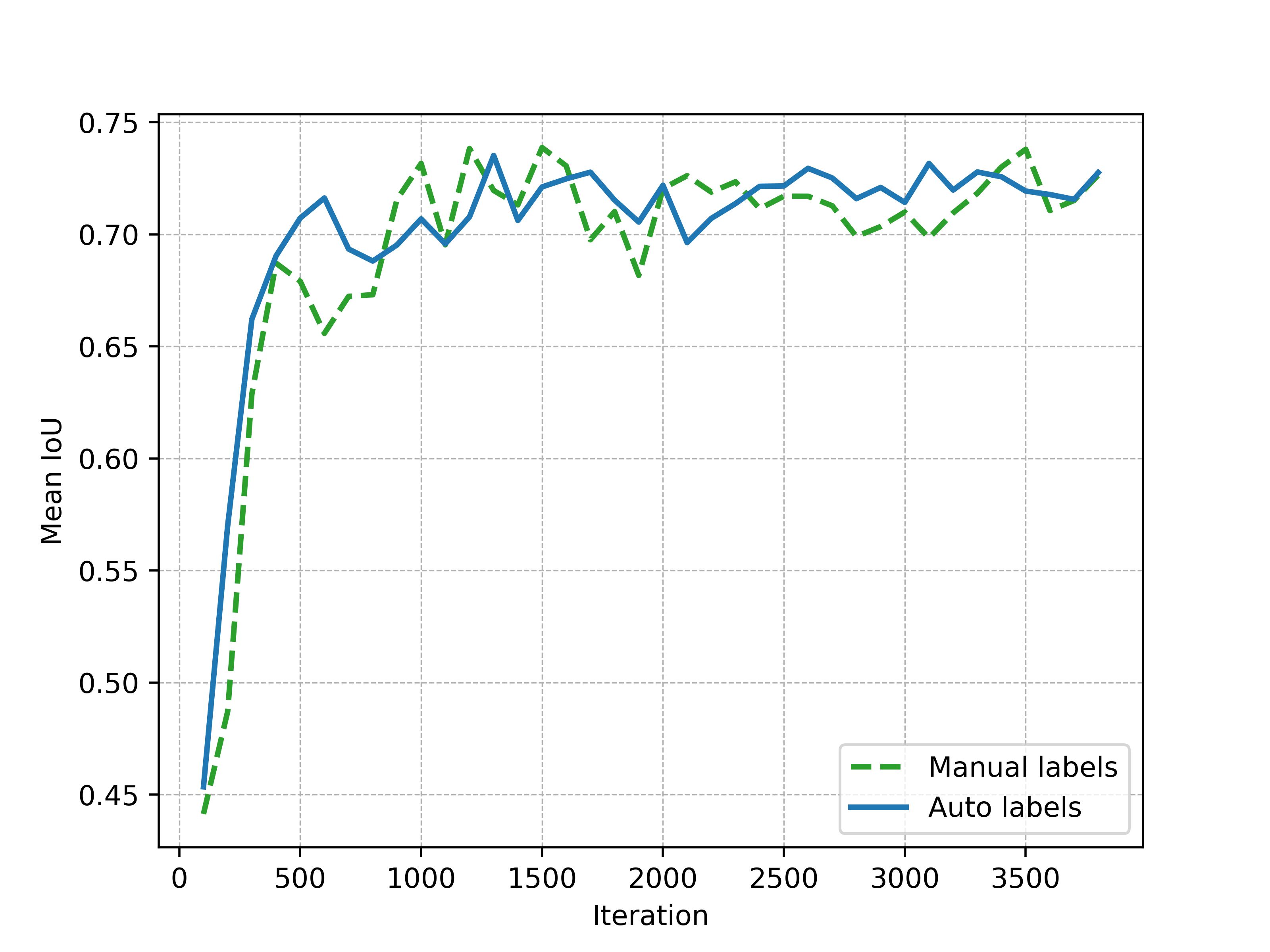}\label{fig_miou_gen}}
	\hfil
	\subfloat[Accuracy]{\includegraphics[width=2.2in]{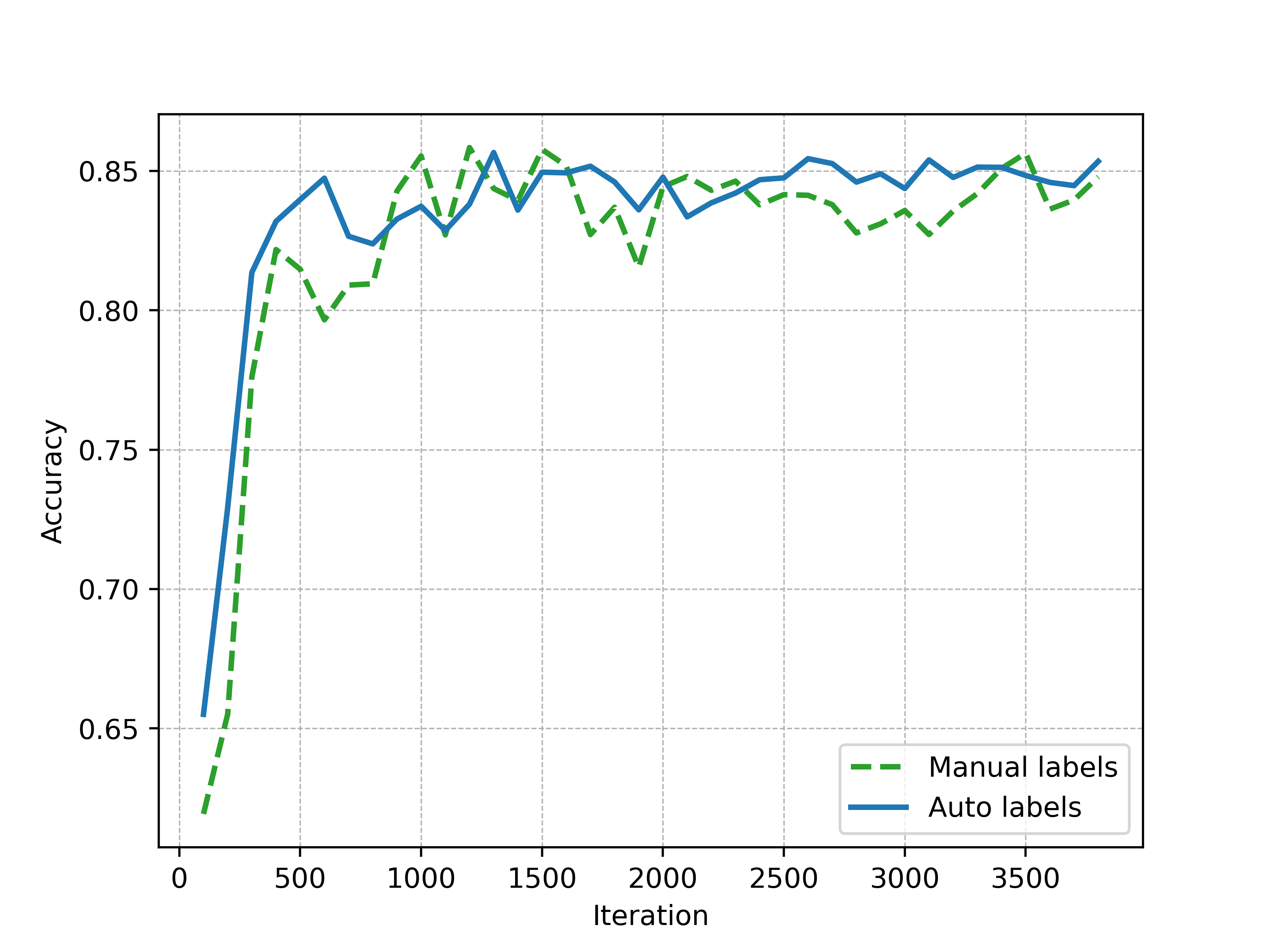}\label{fig_acc_gen}}
	
	\caption{Comparison of model performance trained on auto-annotated versus original labels. The results demonstrate comparable performance, confirming the practical effectiveness of automatic annotation.}
	\label{fig_gen}
\end{figure*}

\begin{figure*}[!t]
	\centering
	\captionsetup[subfloat]{labelformat=empty, position=top}
	
	\subfloat[\fontsize{9pt}{0pt}\selectfont original image]{\includegraphics[width=1.1in]{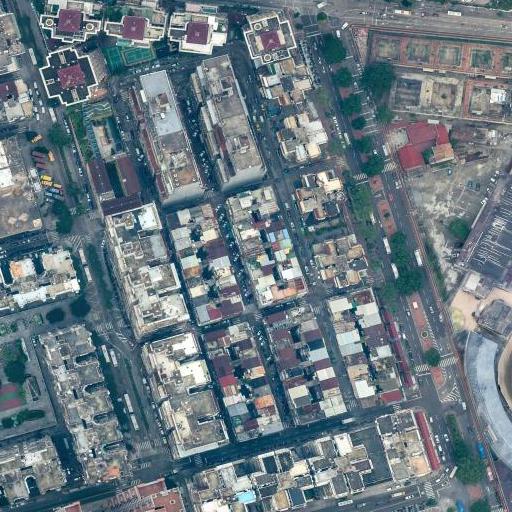}%
		\label{fig_road11}}
	\hfil
	\subfloat[\fontsize{9pt}{0pt}\selectfont manual labels \vspace{2pt}]{\includegraphics[width=1.1in]{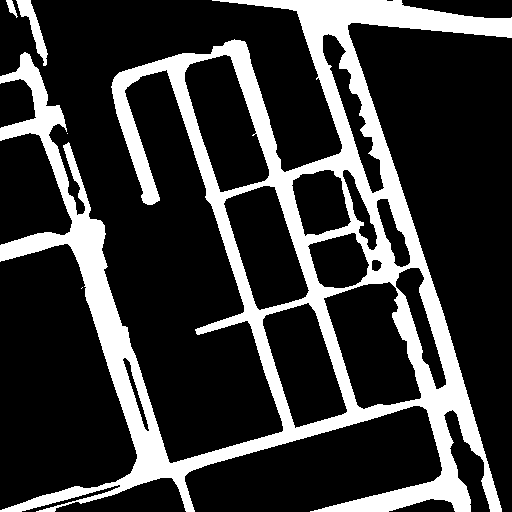}%
		\label{fig_road12}}
	\hfil
	\subfloat[\fontsize{9pt}{0pt}\selectfont auto-labels \vspace{2pt}]{\includegraphics[width=1.1in]{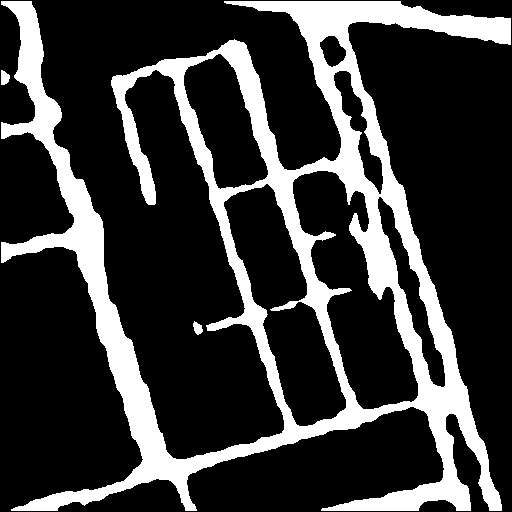}%
		\label{fig_road13}}
	\hfil
	\subfloat[\fontsize{9pt}{0pt}\selectfont original image]{\includegraphics[width=1.1in]{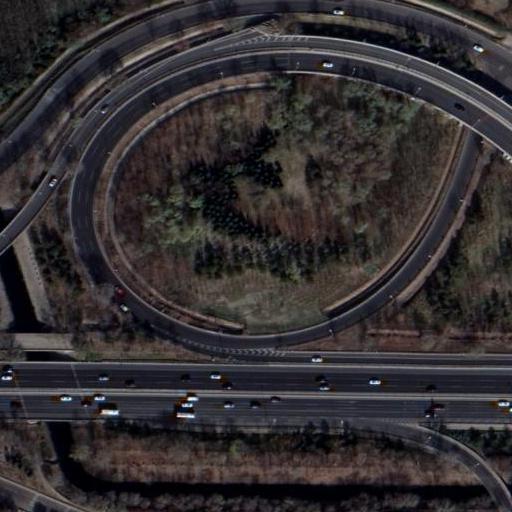}%
		\label{fig_road21}}
	\hfil
	\subfloat[\fontsize{9pt}{1pt}\selectfont manual labels \vspace{2pt}]{\includegraphics[width=1.1in]{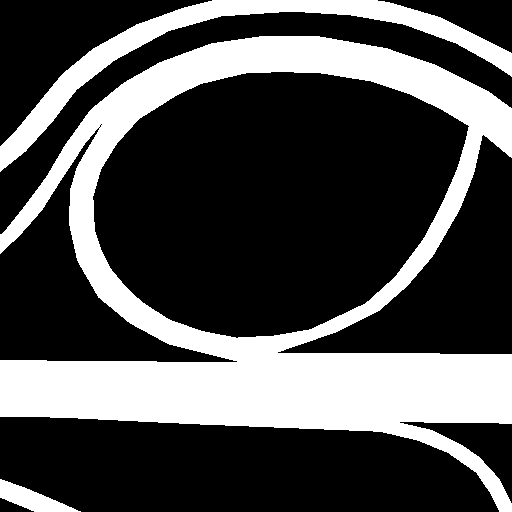}%
		\label{fig_road22}}
	\hfil
	\subfloat[\fontsize{9pt}{1pt}\selectfont auto-labels \vspace{2pt}]{\includegraphics[width=1.1in]{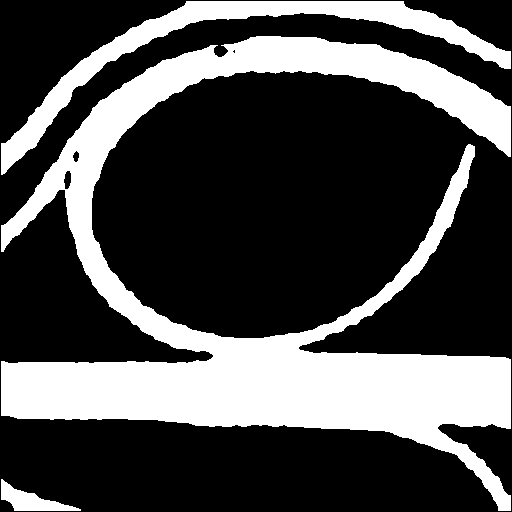}%
		\label{fig_road23}}
	
	\vspace{-20pt}
	
	\subfloat[]{\includegraphics[width=1.1in]{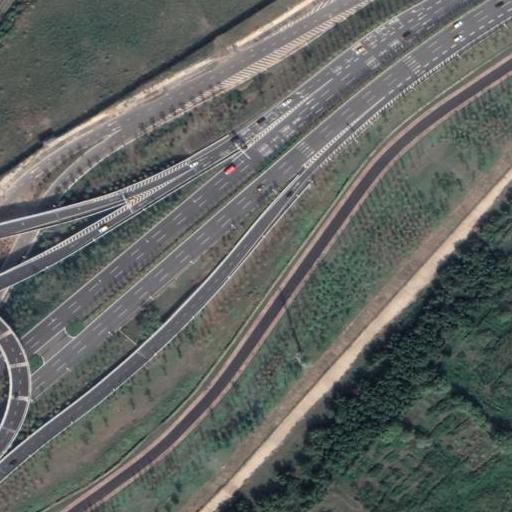}%
		\label{fig_road31}}
	\hfil
	\subfloat[]{\includegraphics[width=1.1in]{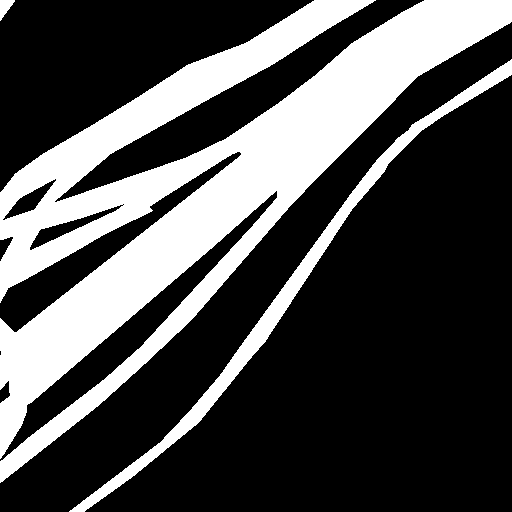}%
		\label{fig_road32}}
	\hfil
	\subfloat[]{\includegraphics[width=1.1in]{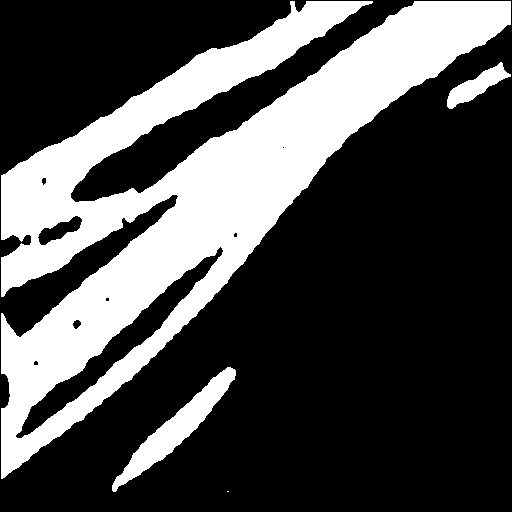}%
		\label{fig_road33}}
	\hfil
	\subfloat[]{\includegraphics[width=1.1in]{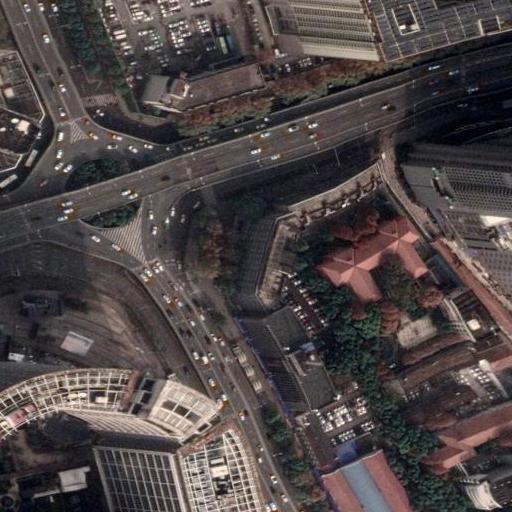}%
		\label{fig_road41}}
	\hfil
	\subfloat[]{\includegraphics[width=1.1in]{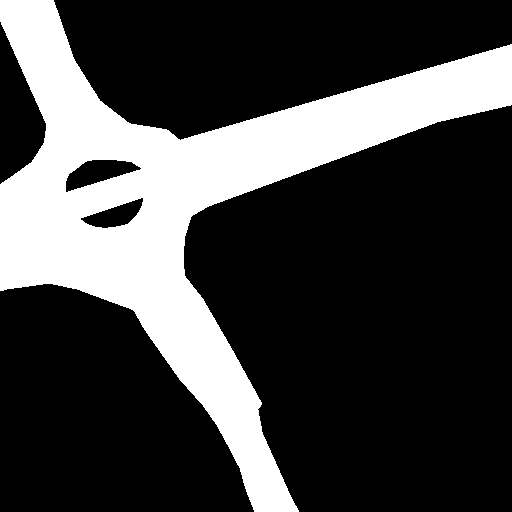}%
		\label{fig_road42}}
	\hfil
	\subfloat[]{\includegraphics[width=1.1in]{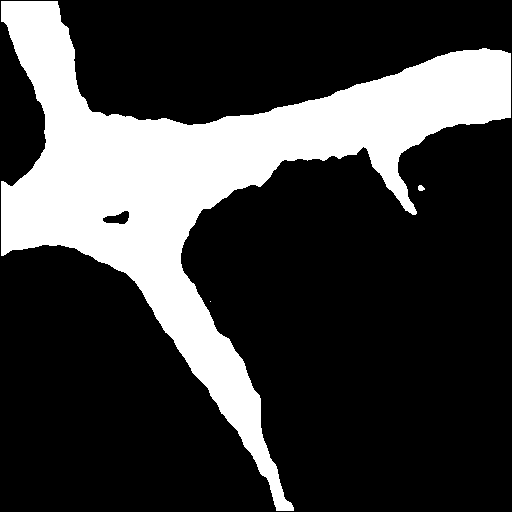}%
		\label{fig_road43}}
	
	\caption{Visual Comparison of Auto-Annotated and Manually Labeled Data. The auto-annotated labels exhibit comparable quality to manual annotations, effectively capturing key structures and fine boundaries in high-resolution remote sensing images.}
	\label{fig_op}
\end{figure*}

\subsection{Automatic Annotation Performance Evaluation}
To further evaluate the generalization capability of the proposed model, particularly its potential in automatic annotation tasks, we design a new set of experiments to simulate a typical cross dataset annotation scenario. Specifically, the model is fine-tuned on the training set of the PALSAR dataset for 200 epochs, and then used to automatically generate labels for the training set of the Sentinel dataset.

\subsubsection{Qualitative Analysis}
To visually assess the quality of automatically generated labels in the Sentinel dataset, we conducted a comparative analysis through visual inspection. As shown in Fig.~\ref{fig_Generalization}, the automatically generated labels demonstrate superior quality compared to the original annotations, characterized by clearer boundaries, finer details, and improved representation of target regions.

\subsubsection{Quantitative Analysis}
To fairly assess the impact of different labels on model performance, we use the same test set across all comparisons—the original test set of the Sentinel dataset. Two models are trained separately: one using the automatically generated labels and the other using the original labels. By comparing their performance on the same test set, we can effectively evaluate the quality and practicality of the automatically generated annotations. This experiment helps analyze the model's ability to adapt to distribution shifts and confirms its potential for cross-dataset automatic annotation tasks.

To ensure consistent evaluation, we further adopt two mainstream semantic segmentation models: DeepLabv3+ (with ResNet50 as the backbone) and U-Net. All models are initialized with the same pretrained weights and trained under identical hyperparameter settings for 20 epochs. Performance on the shared test set is compared to assess model adaptability in cross-domain scenarios and validate the feasibility and effectiveness of the proposed framework for automatic annotation.

\begin{table}[!t]
	\caption{Performance Comparison Between Models Trained on Manual and Auto-Annotated Labels (Sentinel Test Set)}
	\label{tab:generalization_quality}
	\centering
	\renewcommand{\arraystretch}{1.2}
	\begin{tabular}{c c c c c}
		\hline
		\textbf{Test Set} & \textbf{Model} & \textbf{Train Set} & \textbf{ACC} & \textbf{mIoU} \\
		\hline
		\multirow{2}{*}{Original} 
		& DeepLabv3+ & Manual Labels & 0.8577 & 0.7387 \\
		& DeepLabv3+ & Auto Labels   & 0.8566 & 0.7352 \\
		\hline
		\multirow{2}{*}{Original} 
		& U-Net      & Manual Labels & 0.8781 & 0.7672 \\
		& U-Net      & Auto Labels   & 0.8722 & 0.7649 \\
		\hline
	\end{tabular}
\end{table}

As shown in Table~\ref{tab:generalization_quality} and Fig.~\ref{fig_Generalization}, the models trained with auto-annotated labels achieved performance comparable to those trained with original labels on the Sentinel dataset, further confirming the feasibility and effectiveness of the proposed framework in remote sensing applications.

\subsection{Automatic Annotation on Very High-Resolution Remote Sensing Data}
To further validate the applicability of the proposed automatic annotation framework in optical remote sensing scenarios, we conducted an additional set of experiments. Unlike the previous SAR-based experiments, this evaluation focuses on the CHN6-CUG Road Dataset, a high-resolution optical remote sensing dataset. This setup allows us to assess the generalization capability and practical value of our method under different sensor types and stricter annotation standards.

To simulate a realistic ``low-cost few-shot annotation + large-scale automatic annotation" workflow, only the first 30\% of the training data from the dataset was used to fine-tune the model. The fine-tuned model then automatically annotated the remaining training data to generate labels. We trained two independent models: one using the original labels and the other using the automatically generated labels, and evaluated their performance on a shared test set.

This experiment aims to examine whether the proposed framework can efficiently annotate large-scale data with limited high-quality supervision and provide reliable guidance for downstream model training under resource-constrained conditions.

\begin{table}[!t]
	\caption{Performance Comparison Between Models Trained on Manual and Auto-Annotated Labels (CHN6-CUG Road Dataset)}
	\label{tab:op}
	\centering
	\renewcommand{\arraystretch}{1.2}
	\begin{tabular}{c c c c c}
		\hline
		\textbf{Test Set} & \textbf{Model} & \textbf{Train Set} & \textbf{ACC} & \textbf{mIoU} \\
		\hline
		\multirow{2}{*}{Original} 
		& DeepLabv3+ & Manual Labels & 0.9581 & 0.6865 \\
		& DeepLabv3+ & Auto Labels   & 0.9548 & 0.6561 \\
		\hline
		\multirow{2}{*}{Original} 
		& U-Net      & Manual Labels & 0.9556 & 0.6548 \\
		& U-Net      & Auto Labels   & 0.9485 & 0.6379 \\
		\hline
	\end{tabular}
\end{table}

As shown in Table~\ref{tab:op} and Fig.~\ref{fig_op}, models trained on automatically annotated data exhibit slightly lower performance compared to those trained on manual labels. However, the results remain practically valuable, offering a cost-effective solution for real-world applications.

\subsection{Application of the Label Quality Evaluator}

To evaluate the effectiveness of the proposed Label Quality Evaluator (LQE), we conducted an additional experiment on the CHN6-CUG Road dataset. Building upon earlier experiments that compared manually annotated labels and unfiltered auto-generated labels, this study introduced a third setting—auto-generated labels filtered by LQE.

As shown in Table~\ref{tab:lqe_results}, models trained with unfiltered auto-generated labels exhibited a clear decline in both accuracy (ACC) and mean Intersection over Union (mIoU) compared to those trained with manual annotations. This confirms that label noise can negatively affect segmentation performance. However, LQE was applied to screen the auto-generated labels, and model performance improved significantly. In particular, mIoU showed a notable increase. 

To further explore the influence of the Iterative Enhancement hyperparameter, we also tested different loop numbers in the iterative process. As shown in Table~\ref{tab:iterative_enhancement}, we observed that the number of loops affected model performance. Considering both performance and time cost, we recommend using 2 or 3 iterations as a balanced choice for optimal results.

These results validate the utility of the Label Quality Evaluator. LQE  distinguishes high-quality labels from noisy ones, allowing the model to learn from more reliable supervision. This makes LQE a valuable component in scalable annotation pipelines, especially under limited annotation resources.

\begin{table}[!t]
	\caption{Experimental Results Evaluating the Effectiveness of the Label Quality Evaluator (LQE)}
	\label{tab:lqe_results}
	\centering
	\renewcommand{\arraystretch}{1.2}
	\begin{tabular}{c c c c c}
		\hline
		\textbf{Test Set} & \textbf{Model} & \textbf{Train Set} & \textbf{ACC} & \textbf{mIoU} \\
		\hline
		\multirow{3}{*}{Original}
		& DeepLabv3+ & Manual Labels        & 0.9581 & 0.6865 \\
		& DeepLabv3+ & Auto Labels          & 0.9548 & 0.6561 \\
		& DeepLabv3+ & Auto Labels (LQE)    & 0.9572 & 0.6705 \\
		\hline
		\multirow{3}{*}{Original}
		& U-Net      & Manual Labels        & 0.9556 & 0.6548 \\
		& U-Net      & Auto Labels          & 0.9485 & 0.6379 \\
		& U-Net      & Auto Labels (LQE)    & 0.9532 & 0.6469 \\
		\hline
	\end{tabular}
\end{table}

\begin{table}[!t]
	\caption{Experimental Results of Iterative Enhancement with Different Loop Numbers}
	\label{tab:iterative_enhancement}
	\centering
	\renewcommand{\arraystretch}{1.2}
	\begin{tabular}{c c c}
		\hline
		\textbf{Loop Numbers} & \textbf{ACC} & \textbf{mIoU} \\
		\hline
		0 & 0.9548 & 0.6561 \\
		1 & 0.9575 & 0.6632 \\
		2 & 0.9556 & 0.6686 \\
		3 & 0.9572 & 0.6705 \\
		4 & 0.9569 & 0.6697 \\
		\hline
	\end{tabular}
\end{table}

\subsection{Ablation Study: Effect of the EAM}
To investigate the impact of the EAM on remote sensing image segmentation performance, our work conducted an ablation study. By comparing model configurations with and without the EAM, our work evaluated differences in both qualitative visualization and quantitative metrics to assess its contribution to improving segmentation accuracy.
\subsubsection{Qualitative Analysis}
\begin{figure}[!t]
	\centering
	\subfloat[]{\includegraphics[width=1.1in]{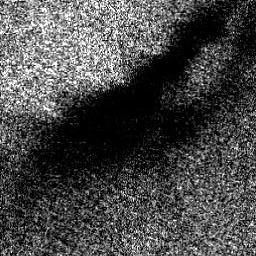}%
		\label{fig_pict1}}
	\hfil
	\subfloat[]{\includegraphics[width=1.1in]{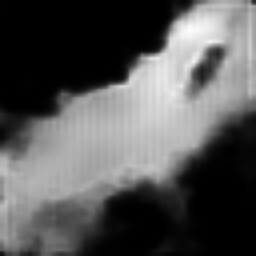}%
		\label{fig_pict2}}
	\hfil
	\subfloat[]{\includegraphics[width=1.1in]{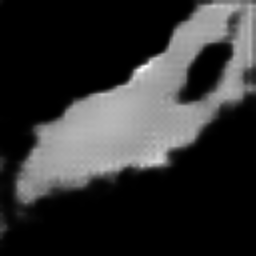}%
		\label{fig_pict3}}
	
	\vspace{-9pt}
	
	\subfloat[]{\includegraphics[width=1.1in]{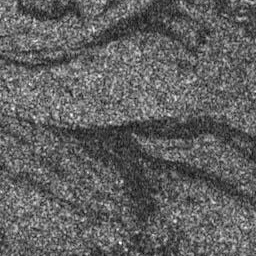}%
		\label{fig_pict4}}
	\hfil
	\subfloat[]{\includegraphics[width=1.1in]{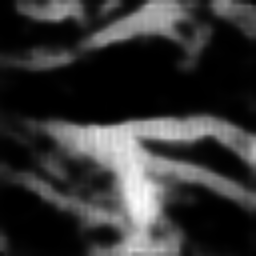}%
		\label{fig_pict5}}
	\hfil
	\subfloat[]{\includegraphics[width=1.1in]{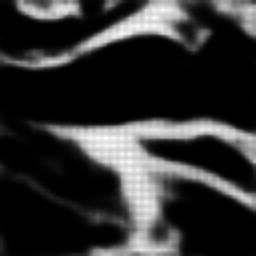}%
		\label{fig_pict6}}

	\caption{Impact of the EAM on Segmentation Results. (b) Without EAM, (c) With EAM; (e) Without EAM, (f) With EAM.}
	\label{fig_edge_attention}
\end{figure}

Fig. \ref{fig_edge_attention} presents the segmentation results obtained with (left) and without (right) the EAM.

Experimental results show that the EAM significantly improves segmentation accuracy along object boundaries. In regions with low contrast or blurred edges, this module captures finer details, generating smoother and more continuous segmentation boundaries while reducing jagged or fragmented edges. In contrast, models without the EAM perform worse in these areas, often distorting shapes and creating ambiguous boundaries, which lowers segmentation quality.

However, introducing the EAM may lead to minor information loss in small-scale regions, causing some small oil spill areas to be overlooked. From an annotation perspective, incorporating edge attention is more suitable for generating segmentation labels, especially in scenarios where clear boundary delineation is crucial.

\subsubsection{Quantitative Analysis}
To further evaluate the contribution of EAM, we constructed two different test sets by enabling and disabling EAM, and compared them with the baseline performance on TEST-HQ. The results are shown in Table~\ref{tab:eam_ablation}.

\begin{table*}[!t]
	\caption{Impact of the EAM on Model Performance (Compared with TEST-HQ Baseline)}
	\label{tab:eam_ablation}
	\centering
	\renewcommand{\arraystretch}{1.2}
	\setlength{\tabcolsep}{6pt}
	{
		\begin{tabular}{c c c c c c c}
			\hline
			\textbf{Test Model} & \textbf{Train Set} & \textbf{Test Set} & \textbf{mIoU} & \textbf{\(\Delta\) from HQ (mIoU)} & \textbf{ACC} & \textbf{\(\Delta\) from HQ (ACC)} \\
			\hline
			\multirow{3}{*}{DeepLabv3+}
			& Original & Test-HQ      & 0.7815 & -        & 0.9283 & - \\
			& Original & Test-NoEMA & 0.8126 & 0.0311   & 0.9391 & 0.0108 \\
			& Original & Test-EMA  & 0.7877 & \textbf{0.0062}  & 0.9328 & \textbf{0.0045} \\
			\hline
		\end{tabular}
	}
\end{table*}

As shown in Table~\ref{tab:eam_ablation}, the test set generated with the EAM produces evaluation results that more closely align with those on the finely annotated high-quality Test-HQ dataset, compared to the version generated without EAM. This suggests that EAM contributes to the generation of higher-quality labels. By incorporating edge-aware constraints during label generation, the model better preserves fine boundary structures, leading to more precise and reliable automatic annotations.

\subsection{Efficiency Evaluation}
To evaluate computational efficiency, we compare the training time per epoch across different deep learning segmentation models. We conduct all experiments on an NVIDIA 4090D GPU using a training dataset of 3,101 images. The results are shown in Table~\ref{tab:efficiency_comparison}.

\begin{table}[!t]
	\caption{Comparison of Single Epoch Training Time Across Different Segmentation Models (Efficiency Evaluation)}
	\label{tab:efficiency_comparison}
	\centering
	\renewcommand{\arraystretch}{1.2}
	\begin{tabular}{c c}
		\hline
		\textbf{Model} & \textbf{Epoch Time (s)} \\
		\hline
		DeepLabv3+ ResNet50  & 48 \\
		DeepLabv3+ ResNet101 & 65 \\
		DeepLabv3+ MobileNet & 29 \\
		LOGCAN++ & 24 \\
		Ours & 26 \\
		\hline
	\end{tabular}
		\vspace{-3mm}
\end{table}

Experimental results indicate that our model achieves significantly lower training time per epoch compared to other models. Specifically, it reduces training time by approximately 45.8\% compared to DeepLabv3+ with ResNet50 and 60\% compared to DeepLabv3+ with ResNet101. Moreover, the training time of our model is comparable to that of lightweight architectures such as LOGCAN++ and DeepLabv3+ with MobileNet, demonstrating that it maintains high computational efficiency while preserving segmentation accuracy.

\subsection{Discussion}
In practical remote sensing segmentation tasks, obtaining large-scale, high-quality pixel-level annotations is often infeasible, especially for new regions, sensors, or task types with scarce labeled data. SAM2-ELNet provides an efficient way to generate new training data from limited annotations through automatic labeling and label enhancement. It is particularly suitable for the following scenarios:
\begin{enumerate}
	\item \textbf{Post-processing of coarse annotations.} Given existing coarse-grained labels, SAM2-ELNet can enhance boundary accuracy and structural completeness, making the labels more suitable for refined model training or evaluation.
	
	\item \textbf{Low-shot annotation-driven automation.} When only a small amount of annotated data is available at the early stage of a project, the model can be fine-tuned with this data and then used to automatically label the entire image set, rapidly generating full-coverage labels for pretraining or iterative refinement.
	
	\item \textbf{Rapid cross-region deployment.} When adapting a trained model to new regions or image sources, SAM2-ELNet can be quickly fine-tuned with a small number of local samples, reducing re-annotation costs and accelerating deployment.
	
\end{enumerate}

The framework requires minimal computational resources for fine-tuning, and in some cases, its automatically generated labels can partially replace manual annotations. This makes it suitable for remote sensing projects with limited resources and high annotation costs. For targets with vague boundaries or ambiguous semantics, manual verification may still be necessary. Future enhancements to the label quality evaluator could further improve reliability.

\section{Conclusion}
This paper proposes SAM2-ELNet, a label enhancement and automatic annotation framework tailored for remote sensing under resource constraints. It freezes the SAM2 backbone and fine-tunes lightweight modules to quickly adapt to diverse segmentation tasks with limited annotations. To improve label quality, it incorporates an EAM for boundary construction and a label quality evaluator for filtering and feedback correction.

Experiments on the Deep-SAR Oil Spill and CHN6-CUG Road datasets demonstrate that SAM2-ELNet achieves both effective label enhancement and reliable automatic annotation. With only 30\% of the annotated data used for fine-tuning, the model automatically labeled the entire dataset. Training on these generated labels yielded slightly lower accuracy than the model trained with the full set of original manual annotations, but still confirmed the practicality of the automatic annotation mechanism.

Future work includes extending SAM2-ELNet to multi-class and multi-modal annotation by incorporating textual semantics for vision-language fusion. The label quality evaluator may also be redesigned as a discriminator in an adversarial framework to enable self-improving learning. Research on automatic annotation is expected to provide an efficient and scalable solution for large-scale intelligent interpretation in remote sensing.

%\section*{Acknowledgments}

\bibliographystyle{IEEEtran}
\bibliography{references}

\begin{thebibliography}{10}
\providecommand{\url}[1]{#1}
\csname url@samestyle\endcsname
\providecommand{\newblock}{\relax}
\providecommand{\bibinfo}[2]{#2}
\providecommand{\BIBentrySTDinterwordspacing}{\spaceskip=0pt\relax}
\providecommand{\BIBentryALTinterwordstretchfactor}{4}
\providecommand{\BIBentryALTinterwordspacing}{\spaceskip=\fontdimen2\font plus
\BIBentryALTinterwordstretchfactor\fontdimen3\font minus
  \fontdimen4\font\relax}
\providecommand{\BIBforeignlanguage}[2]{{%
\expandafter\ifx\csname l@#1\endcsname\relax
\typeout{** WARNING: IEEEtran.bst: No hyphenation pattern has been}%
\typeout{** loaded for the language `#1'. Using the pattern for}%
\typeout{** the default language instead.}%
\else
\language=\csname l@#1\endcsname
\fi
#2}}
\providecommand{\BIBdecl}{\relax}
\BIBdecl

\bibitem{ref1}
J.~A. Richards, \emph{Remote Sensing Digital Image Analysis - An
  Introduction}.\hskip 1em plus 0.5em minus 0.4em\relax Springer Berlin
  Heidelberg, 2012.

\bibitem{ref2}
A.~F.~H. Goetz, G.~Vane, J.~Solomon \emph{et~al.}, ``Imaging spectrometry for
  earth remote sensing,'' \emph{Science}, vol. 228, no. 4704, pp. 1147--1153,
  1985.

\bibitem{ref3}
A.~F.~H. Goetz, ``Three decades of hyperspectral remote sensing of the earth: A
  personal view,'' \emph{Remote Sens. Environ.}, vol. 113, pp. S5--S16, 2009.

\bibitem{ref4}
Z.~L. Li, H.~Wu, N.~Wang \emph{et~al.}, ``Land surface emissivity retrieval
  from satellite data,'' \emph{Int. J. Remote Sens.}, vol.~34, no. 9--10, pp.
  3084--3127, 2013.

\bibitem{ref5}
A.~Moreira, P.~Prats-Iraola, M.~Younis \emph{et~al.}, ``A tutorial on synthetic
  aperture radar,'' \emph{IEEE Geosci. Remote Sens. Mag.}, vol.~3, pp. 6--43,
  2013.

\bibitem{ref37}
P.~O. Gislason, J.~A. Benediktsson, and J.~R. Sveinsson, ``Random forests for
  land cover classification,'' \emph{Pattern Recognit. Lett.}, vol.~27, no.~4,
  pp. 294--300, Mar. 2006.

\bibitem{ref38}
X.~Lu, J.~Zhang, T.~Li, and G.~Zhang, ``Synergetic classification of long-wave
  infrared hyperspectral and visible images,'' \emph{IEEE J. Sel. Topics Appl.
  Earth Observ. Remote Sens.}, vol.~8, no.~7, pp. 3546--3557, Jul. 2015.

\bibitem{ref16}
L.~Gao \emph{et~al.}, ``Subspace-based support vector machines for
  hyperspectral image classification,'' \emph{IEEE Geosci. Remote Sens. Lett.},
  vol.~12, no.~2, pp. 349--353, Feb. 2015.

\bibitem{ref17}
P.~Kr{\"a}henb{\"u}hl and V.~Koltun, ``Efficient inference in fully connected
  crfs with gaussian edge potentials,'' in \emph{Proc. Adv. Neural Inf.
  Process. Syst. (NeurIPS)}, vol.~24, 2011, pp. 1--9.

\bibitem{ref49}
X.~Ma, X.~Zhang, M.-O. Pun, and M.~Liu, ``A multilevel multimodal fusion
  transformer for remote sensing semantic segmentation,'' \emph{IEEE Trans.
  Geosci. Remote Sens.}, vol.~62, pp. 1--15, 2024, art no. 5403215.

\bibitem{ref50}
C.~Lin, J.~Zhou, L.~Yin, R.~Bouabid, D.~Mulla, E.~Benami, and Z.~Jin,
  ``Sub-national scale mapping of individual olive trees integrating earth
  observation and deep learning,'' \emph{ISPRS J. Photogramm. Remote Sens.},
  vol. 217, pp. 18--31, 2024.

\bibitem{ref51}
W.~Li, K.~Chen, and Z.~Shi, ``Geographical supervision correction for remote
  sensing representation learning,'' \emph{IEEE Trans. Geosci. Remote Sens.},
  vol.~60, pp. 1--20, 2022.

\bibitem{ref52}
K.~Li, G.~Wan, G.~Cheng, L.~Meng, and J.~Han, ``Object detection in optical
  remote sensing images: A survey and a new benchmark,'' \emph{ISPRS J.
  Photogramm. Remote Sens.}, vol. 159, pp. 296--307, Jan. 2020.

\bibitem{ref53}
C.~Shi, X.~Zhang, L.~Wang, and Z.~Jin, ``A lightweight convolution neural
  network based on joint features for remote sensing scene image
  classification,'' \emph{Int. J. Remote Sens.}, vol.~44, no.~21, pp.
  6615--6641, Nov. 2023.

\bibitem{ref67}
L.~Yin, R.~Ghosh, C.~Lin \emph{et~al.}, ``Mapping smallholder cashew
  plantations to inform sustainable tree crop expansion in benin,''
  \emph{Remote Sens. Environ.}, vol. 295, p. 113695, 2023.

\bibitem{ref6}
L.~Peng, T.~Wei, X.~Chen, X.~Chen, R.~Sun, L.~Wan, J.~Chen, and X.~Zhu,
  ``Human-annotated label noise and their impact on convnets for remote sensing
  image scene classification,'' \emph{IEEE Journal of Selected Topics in
  Applied Earth Observations and Remote Sensing}, vol.~18, pp. 1500--1514,
  2025.

\bibitem{ref7}
A.~Dosovitskiy, L.~Beyer, A.~Kolesnikov, D.~Weissenborn, X.~Zhai,
  T.~Unterthiner, M.~Dehghani, M.~Minderer, G.~Heigold, S.~Gelly, J.~Uszkoreit,
  and N.~Houlsby, ``An image is worth 16x16 words: Transformers for image
  recognition at scale,'' 2021, \textit{arXiv:2010.11929}.

\bibitem{ref9}
Z.~Wu, Y.~Xiong, S.~X. Yu, and D.~Lin, ``Unsupervised feature learning via
  non-parametric instance discrimination,'' in \emph{Proc. IEEE/CVF Conf.
  Comput. Vis. Pattern Recognit. (CVPR)}, Jun. 2018, pp. 3733--3742.

\bibitem{ref11}
T.~Chen, S.~Kornblith, M.~Norouzi, and G.~Hinton, ``A simple framework for
  contrastive learning of visual representations,'' in \emph{Proc. Int. Conf.
  Mach. Learn. (ICML)}, 2020, pp. 1597--1607.

\bibitem{ref10}
K.~He, H.~Fan, Y.~Wu, S.~Xie, and R.~B. Girshick, ``Momentum contrast for
  unsupervised visual representation learning,'' in \emph{Proc. IEEE/CVF Conf.
  Comput. Vis. Pattern Recognit. (CVPR)}, 2020, pp. 9729--9738.

\bibitem{ref54}
J.-B. Grill, F.~Strub, F.~Altché, C.~Tallec, P.~H. Richemond, E.~Buchatskaya,
  C.~Doersch, B.~A. Pires, Z.~D. Guo, M.~G. Azar, B.~Piot, K.~Kavukcuoglu,
  R.~Munos, and M.~Valko, ``Bootstrap your own latent: A new approach to
  self-supervised learning,'' in \emph{Proc. Adv. Neural Inf. Process. Syst.
  (NeurIPS)}, vol.~33.\hskip 1em plus 0.5em minus 0.4em\relax Curran Associates
  Inc., 2020, pp. 21\,271--21\,284.

\bibitem{ref8}
M.~Caron \emph{et~al.}, ``Emerging properties in self-supervised vision
  transformers,'' in \emph{Proc. IEEE/CVF Int. Conf. Comput. Vis. (ICCV)},
  Montreal, QC, Canada, 2021, pp. 9630--9640.

\bibitem{ref12}
K.~He, X.~Chen, S.~Xie, Y.~Li, P.~Dollár, and R.~Girshick, ``Masked
  autoencoders are scalable vision learners,'' in \emph{Proc. IEEE/CVF Conf.
  Comput. Vis. Pattern Recognit. (CVPR)}, 2022, pp. 15\,979--15\,988.

\bibitem{ref13}
A.~Kirillov, E.~Mintun, N.~Ravi, H.~Mao, C.~Rolland, L.~Gustafson, T.~Xiao,
  S.~Whitehead, A.~C. Berg, W.~Y. Lo \emph{et~al.}, ``Segment anything,'' in
  \emph{Proc. IEEE/CVF Int. Conf. Comput. Vis. (ICCV)}, 2023, pp. 4015--4026.

\bibitem{ref14}
H.~Mei, G.~P. Ji, Z.~Wei, X.~Yang, X.~Wei, and D.~P. Fan, ``Camouflaged object
  segmentation with distraction mining,'' in \emph{Proc. IEEE/CVF Conf. Comput.
  Vis. Pattern Recognit. (CVPR)}, 2021, pp. 8772--8781.

\bibitem{ref58}
{CVHub}, ``Advanced auto labeling solution with added features,'' GitHub
  repository, 2023, \textit{Online}.
  Available:https://github.com/CVHub520/X-AnyLabeling.

\bibitem{ref36}
M.~Luo, T.~Zhang, S.~Wei \emph{et~al.}, ``Sam-rsis: Progressively adapting sam
  with box prompting to remote sensing image instance segmentation,''
  \emph{IEEE Trans. Geosci. Remote Sens.}, 2024.

\bibitem{ref35}
X.~Ma, Q.~Wu, X.~Zhao, X.~Zhang, M.-O. Pun, and B.~Huang, ``Sam-assisted remote
  sensing imagery semantic segmentation with object and boundary constraints,''
  \emph{IEEE Trans. Geosci. Remote Sens.}, vol.~62, pp. 1--16, 2024, art. no.
  5636916.

\bibitem{ref59}
K.~Chen, C.~Liu, H.~Chen, H.~Zhang, W.~Li, Z.~Zou, and Z.~Shi, ``Rsprompter:
  Learning to prompt for remote sensing instance segmentation based on visual
  foundation model,'' \emph{IEEE Trans. Geosci. Remote Sens.}, vol.~62, pp.
  1--17, 2024.

\bibitem{ref18}
L.~Yang, H.~Chen, A.~Yang, and J.~Li, ``Easyseg: An error-aware domain
  adaptation framework for remote sensing imagery semantic segmentation via
  interactive learning and active learning,'' \emph{IEEE Trans. Geosci. Remote
  Sens.}, vol.~62, pp. 1--18, 2024, art. no. 4407518.

\bibitem{ref19}
C.~Liu, C.~M. Albrecht, Y.~Wang, and X.~X. Zhu, ``Task specific pretraining
  with noisy labels for remote sensing image segmentation,'' in \emph{Proc.
  IEEE Int. Geosci. Remote Sens. Symp. (IGARSS)}, Athens, Greece, 2024, pp.
  7040--7044.

\bibitem{ref63}
H.~Yuan, X.~Li, T.~Zhang \emph{et~al.}, ``Sa2va: Marrying sam2 with llava for
  dense grounded understanding of images and videos,'' 2025,
  \textit{arXiv:2501.04001}.

\bibitem{ref20}
W.~Wu, M.~S. Wong, X.~Yu, G.~Shi, C.~Y.~T. Kwok, and K.~Zou, ``Compositional
  oil spill detection based on object detector and adapted segment anything
  model from sar images,'' \emph{IEEE Geosci. Remote Sens. Lett.}, vol.~21, pp.
  1--5, 2024, art. no. 4007505.

\bibitem{ref21}
H.~Wu, Z.~Du, D.~Zhong, Y.~Wang, and C.~Tao, ``Fsvlm: A vision-language model
  for remote sensing farmland segmentation,'' \emph{IEEE Trans. Geosci. Remote
  Sens.}, vol.~63, pp. 1--13, 2025, art. no. 4402813.

\bibitem{ref22}
J.~Li, Y.~Shen, and W.~Ye, ``Interactive segmentation method utilizing spectral
  and contour features: A case study on a dual-task remote sensing water body
  dataset,'' \emph{IEEE Geosci. Remote Sens. Lett.}, vol.~21, pp. 1--5, 2024,
  art. no. 1503705.

\bibitem{ref23}
W.~Han \emph{et~al.}, ``Dual-model collaboration consistency semi-supervised
  learning for few-shot lithology interpretation,'' \emph{IEEE Trans. Geosci.
  Remote Sens.}, vol.~62, pp. 1--14, 2024, art. no. 4514114.

\bibitem{ref24}
C.~Liu, C.~M. Albrecht, Y.~Wang, Q.~Li, and X.~X. Zhu, ``Aio2: Online
  correction of object labels for deep learning with incomplete annotation in
  remote sensing image segmentation,'' \emph{IEEE Trans. Geosci. Remote Sens.},
  vol.~62, pp. 1--17, 2024, art. no. 5613917.

\bibitem{ref25}
J.~Shi \emph{et~al.}, ``Fine object change detection based on vector boundary
  and deep learning with high-resolution remote sensing images,'' \emph{IEEE J.
  Sel. Topics Appl. Earth Observ. Remote Sens.}, vol.~15, pp. 4094--4103, 2022.

\bibitem{ref26}
J.~E. Gallagher, A.~Gogia, and E.~J. Oughton, ``A multispectral automated
  transfer technique (matt) for machine-driven image labeling utilizing the
  segment anything model (sam),'' \emph{IEEE Access}, vol.~13, pp. 4499--4516,
  2025.

\bibitem{ref27}
C.~Zhang, P.~Marfatia, H.~Farhan, L.~Di, L.~Lin, H.~Zhao \emph{et~al.},
  ``Enhancing usda nass cropland data layer with segment anything model,'' in
  \emph{Proc. 11th Int. Conf. Agro-Geoinformatics}, 2023, pp. 1--5.

\bibitem{ref28}
B.~Xue, H.~Cheng, Q.~Yang, Y.~Wang, and X.~He, ``Adapting segment anything
  model to aerial land cover classification with low-rank adaptation,''
  \emph{IEEE Geosci. Remote Sens. Lett.}, vol.~21, pp. 1--5, 2024.

\bibitem{ref29}
Q.~Liu and J.~Ma, ``Foundation models for geophysics: Review and perspective,''
  2024, \textit{arXiv:2406.03163}.

\bibitem{ref32}
B.~M. Martins and G.~Salgado, ``Application of the sam (segment anything model)
  algorithm to cbers-4a/wpm remote sensing images for the identification of
  urban areas in s{\~a}o sebasti{\~a}o/sp, brazil,'' in \emph{Brazilian
  Symposium on GeoInformatics}, 2023.

\bibitem{ref33}
C.~Lee, S.~Soedarmadji, M.~Anderson, A.~J. Clark, and S.-J. Chung, ``Semantics
  from space: Satellite-guided thermal semantic segmentation annotation for
  aerial field robots,'' 2024, \textit{arXiv:2403.14056}.

\bibitem{ref34}
S.~Ren, F.~Luzi, S.~Lahrichi, K.~Kassaw, L.~M. Collins, K.~Bradbury
  \emph{et~al.}, ``Segment anything from space?'' in \emph{Proc. IEEE/CVF
  Winter Conf. Appl. Comput. Vis. (WACV)}, Jan. 2024, pp. 8340--8350.

\bibitem{ref68}
X.~Zhou, F.~Liang, L.~Chen, Liu \emph{et~al.}, ``Mesam: Multiscale enhanced
  segment anything model for optical remote sensing images,'' \emph{IEEE Trans.
  Geosci. Remote Sens.}, vol.~62, pp. 1--15, 2024.

\bibitem{ref15}
X.~Xiong, Z.~Wu, S.~Tan, W.~Li, F.~Tang, Y.~Chen, S.~Li, J.~Ma, and G.~Li,
  ``Sam2-unet: Segment anything 2 makes strong encoder for natural and medical
  image segmentation,'' 2024, \textit{arXiv:2408.08870}.

\bibitem{ref60}
N.~Houlsby, A.~Giurgiu, S.~Jastrzebski, B.~Morrone, Q.~D. Laroussilhe,
  A.~Gesmundo, M.~Attariyan, and S.~Gelly, ``Parameter-efficient transfer
  learning for nlp,'' in \emph{Proc. Int. Conf. Mach. Learn. (ICML)}, 2019, pp.
  2790--2799.

\bibitem{ref61}
Z.~Qiu, Y.~Hu, H.~Li, and J.~Liu, ``Learnable ophthalmology sam,'' 2023,
  \textit{arXiv:2304.13425}.

\bibitem{ref39}
E.~J. Hu, Y.~Shen, P.~Wallis, Z.~Allen-Zhu, Y.~Li, S.~Wang, L.~Wang, and
  W.~Chen, ``Lora: Low-rank adaptation of large language models,'' 2022,
  \textit{arXiv:2106.09685}.

\bibitem{ref40}
C.~Ryali, Y.~T. Hu, D.~Bolya, C.~Wei, H.~Fan, P.~Y. Huang, V.~Aggarwal,
  A.~Chowdhury, O.~Poursaeed, J.~Hoffman, and et~al., ``Hiera: A hierarchical
  vision transformer without the bells-and-whistles,'' in \emph{Proc. Int.
  Conf. Mach. Learn. (ICML)}.\hskip 1em plus 0.5em minus 0.4em\relax PMLR,
  2023, pp. 29\,441--29\,454.

\bibitem{ref55}
S.~Liu, D.~Huang \emph{et~al.}, ``Receptive field block net for accurate and
  fast object detection,'' in \emph{Proc. Eur. Conf. Comput. Vis. (ECCV)},
  2018, pp. 385--400.

\bibitem{ref56}
D.~P. Fan, G.~P. Ji, T.~Zhou, G.~Chen, H.~Fu, J.~Shen, and L.~Shao, ``Pranet:
  Parallel reverse attention network for polyp segmentation,'' in \emph{Proc.
  Med. Image Comput. Comput.-Assist. Interv. (MICCAI)}.\hskip 1em plus 0.5em
  minus 0.4em\relax Springer, 2020, pp. 263--273.

\bibitem{ref57}
D.~Hendrycks and K.~Gimpel, ``Gaussian error linear units (gelus),''
  \emph{arXiv preprint}, vol. arXiv:1606.08415, 2023.

\bibitem{ref46}
Q.~Zhu \emph{et~al.}, ``Oil spill contextual and boundary-supervised detection
  network based on marine sar images,'' \emph{IEEE Trans. Geosci. Remote
  Sens.}, vol.~60, pp. 1--10, 2022, art no. 5213910.

\bibitem{ref64}
Q.~Zhu, Y.~Zhang, L.~Wang \emph{et~al.}, ``A global context-aware and batch
  independent network for road extraction from vhr satellite imagery,''
  \emph{ISPRS J. Photogramm. Remote Sens.}, vol. 175, pp. 353--365, 2021.

\bibitem{ref62}
O.~Ronneberger, P.~Fischer, and T.~Brox, ``U-net: Convolutional networks for
  biomedical image segmentation,'' in \emph{Proc. Med. Image Comput.
  Comput.-Assist. Interv. (MICCAI)}.\hskip 1em plus 0.5em minus 0.4em\relax
  Cham: Springer, 2015, pp. 234--241.

\bibitem{ref66}
X.~Ma, R.~Lian, Z.~Wu \emph{et~al.}, ``Logcan++: Adaptive local-global
  class-aware network for semantic segmentation of remote sensing images,''
  \emph{IEEE Trans. Geosci. Remote Sens.}, vol.~63, pp. 1--16, 2025.

\bibitem{ref47}
L.-C. Chen, G.~Papandreou, F.~Schroff, and H.~Adam, ``Rethinking atrous
  convolution for semantic image segmentation,'' Dec. 2017,
  \textit{arXiv:1706.05587}.

\end{thebibliography}

\vfill

\end{document}